\documentclass{article}

\usepackage[preprint]{neurips_2024}

\author{
  Dylan Xu \\
  Department of Computer Science\\
  University of California, Berkeley\\
  Berkeley, CA 94720 \\
  \texttt{dylanx26@berkeley.edu} \\
  \And
  Juan-Pablo Rivera \\
  Department of Compuer Science \\
  Georgia Tech University \\
  Atlanta, GA 30332 \\
  \texttt{jprivera64@gatech.edu} \\
}

\usepackage[utf8]{inputenc} 
\usepackage[T1]{fontenc}    
\usepackage{hyperref}       
\usepackage{url}            
\usepackage{booktabs}       
\usepackage{amsfonts}       
\usepackage{nicefrac}       
\usepackage{microtype}      
\usepackage{xcolor}         
\usepackage{amsmath}
\usepackage{graphicx}
\usepackage{subcaption}

\newtheorem{definition}{Definition}[section]
\usepackage{verbatim}

\usepackage{float}

\title{Towards Measuring Goal-Directedness in AI Systems}

\begin{document}

\maketitle

\begin{abstract}
    Recent advances in deep learning have brought attention to the possibility of creating advanced, general AI systems that outperform humans across many tasks. However, if these systems pursue unintended goals, there could be catastrophic consequences. A key prerequisite for AI systems pursuing unintended goals is whether they will behave in a coherent and goal-directed manner in the first place, optimizing for some unknown goal; there exists significant research trying to evaluate systems for said behaviors. However, the most rigorous definitions of goal-directedness we currently have are difficult to compute in real-world settings. Drawing upon this previous literature, we explore policy goal-directedness within reinforcement learning (RL) environments. In our findings, we propose a different family of definitions of the goal-directedness of a policy that analyze whether it is well-modeled as near-optimal for many (sparse) reward functions. We operationalize this preliminary definition of goal-directedness and test it in toy Markov decision process (MDP) environments. Furthermore, we explore how goal-directedness could be measured in frontier large-language models (LLMs). Our contribution is a definition of goal-directedness that is simpler and more easily computable in order to approach the question of whether AI systems could pursue dangerous goals. We recommend further exploration of measuring coherence and goal-directedness, based on our findings.

\end{abstract}

\section{Introduction}

In recent years, the field of deep learning has seen remarkable advances in capabilities and generality from training large neural networks on large corpuses of data. For example, transformers (\cite{vaswani2023attention}) have been successful in NLP (\cite{brown2020language}), image generation (\cite{ramesh2021zeroshot}), and protein structure prediction (\cite{JumperAlphafold2021}). Deep learning has also found success in games like Go (\cite{SilverGo2016}) and Starcraft 2 (\cite{Vinyals2019}). These successes have raised the possibility, now being pursued by multiple companies (\cite{openaiagi}, \cite{deepmindagi}), of advanced artificial general intelligences (AGIs) that achieve better than human performance in a wide variety of tasks. 

However, there are concerns that AGIs could generalize poorly to unintended behavior in unforeseen environments. In particular, if an AGI is well-modeled as \textit{goal-directed}, then it could pursue a misaligned goal that could lead to unwanted behavior such as power-seeking (\cite{shah2022goal, langosco2023goal}). A particularly dangerous hypothetical failure mode arising from \textit{goal misgeneralization} is if the AGI learns to \textit{scheme} against or deceive its training process to protect its misaligned goal from being changed (\cite{hubinger2019risksfrom}, \cite{carlsmith2023scheming}). A significant proportion of current AI safety work focuses on detecting and analyzing (e.g. via analogous case studies of model organisms like \cite{hubinger2023model, hubinger2019risksfrom}) scheming-type behavior in AI systems. If an AGI is vastly more intelligent than humans, then goal misgeneralization could lead to the disempowerment of humanity or other worst-case scenarios (\cite{ngo2022deeplearning}, \cite{carlsmith2023scheming}).

Previous work in this area, which we discuss further in Section 2, present a set of conceptual and mathematical definitions for goal-directedness, coherence, and/or agency (\cite{litreview}). However, these definitions are usually only tested in toy settings and are difficult to realistically compute. In this paper, we explore the goal-directedness of a policy as whether it is well-modeled as near-optimal for many sparse reward functions, and give a method to train a goal-directedness classifier in Section 3. We present preliminary experiments in Section 4 with increasing complexity to demonstrate the tractability and results of our method. Our results show that it is possible to separate a dense and sparse reward function, although more robust testing needs to happen. 

Overall, we broadly propose focusing on measuring goal-directedness as a new research direction to help reduce the risk of catastrophic misalignment. We discuss the possible impacts of our research in Section 5. 
\footnote{We also note that the question of defining goal-directedness or agency is relevant in other fields of artificial intelligence like robotics and RL (\cite{Hanheide2010AFF}, \cite{Butlin2022MachineLF}), multiagent systems (\cite{Luck1995AFF}), and philosophy (\cite{Butlin2022MachineLF}, \cite{Dung2024UnderstandingAA}, \cite{Heylighen2022TheMA}). While we focus on goal-directedness as it relates to AI safety in this document, other fields may find our work useful.}

\section{Background and Related Work}

One paper related to our methods is \cite{agentsanddevices} (cited by \cite{shah2022goal}), which uses a variant of Bayesian inverse reinforcement learning (\cite{IRLoriginal}, \cite{ChoiBayesianIRL}) to calculate the posterior probability that a system is well described as an ``agent" versus a ``device". This definition is explicitly based on the intentional stance (\cite{dennettintentional}), which roughly claims that there is no observer-independent truth as to whether a system is or is not a goal-directed agent; we can only describe how \textit{well-modeled} a system is as an agent by noticing behavioral patterns corresponding to things we might call ``beliefs" or ``desires". However, \cite{agentsanddevices} only test their method in small gridworlds, while we attempt experiments in more complex environments. More mathematical equivalences are provided in section \ref{Mathematical equivalences}.


A related concept is that goal-directed models should be able to adapt and generalize to out-of-distribution settings well (\cite{KENTON2023103963}, \cite{turner2022parametrically}). Similar to these definitions, our method focuses on the mechanistic internals of a policy, however our definition also  relies on the intentional stance. Previous literature often models the preferences of a possible agent with a utility or reward function, which is justified theoretically through coherence theorems in decision theory that equate axioms about preferences to expected utility maximization (\cite{vnmtheorem}, \cite{savage54}).

Further details of the relevant background can be found in Section \ref{Further background}.

\section{Methods and Key Definitions}

We now present our mathematical model for measuring the goal-directedness of a policy in an environment. Inspired by the intentional framework (\cite{dennettintentional}), goal-directed AIs are AIs that are ``well-described" as having a goal, which we can formalize as a reward function in the context of sequential decision-making (\cite{agentsanddevices}).

Given some environment with states and transitions, goal-directed policies should be good at \textit{many} reward functions $R(s, a, s’)$, indicating their ability to \textit{generalize} across states in the environment \textit{and} across tasks. (This is a common thread in other goal-directedness definitions, such as \cite{agentsanddevices} and \cite{KENTON2023103963}.) Such a goal-directed policy would need to adapt to new circumstances and likely internally need to select actions based on their consequences, which are core features of goal-directedness. Our definition is similar to (a simplified version of) \cite{kosoydefinition} and \cite{agentsanddevices}; we compare methods in further detail in section \ref{Mathematical equivalences}. We give a full mathematical definition of our model in deterministic Markov decision processes (MDP, \cite{MDPintro}) without preset reward functions in section \ref{Mathematical model}. Briefly speaking, in an MDP our methodology is to generate different sets of policies in increasing order of goal-directedness:

\begin{enumerate}
    \item \textbf{UPS}: Sample a random policy uniformly from the space of all policies.
    \item \textbf{URS}: Sample a random reward function over states or transitions uniformly from the space of all reward functions, then sample a near-optimal policy for that reward function.
    \item \textbf{USS}: Sample a percentage of states or transitions to assign reward to, while assigning near-zero reward to all other states/transitions. This results in a \textit{sparse} reward function over the environment. Theoretically, policies likely to be drawn using USS should be \textit{far-sighted}, or able to consider consequences far in the future across multiple time-steps. We then sample a reward function for your chosen states/transitions, then sample a near-optimal policy. \footnote{Alternatively, one could sample from the space of reward functions with some \textit{simplicity prior} distribution, similar to \cite{kosoydefinition}.}
\end{enumerate}

We choose two sampling strategies, then define the goal-directedness $G(\pi_0)$ of a policy as \textit{the ratio of how likely it is under the more goal-directed sampling strategy to how likely it is under the less goal-directed sampling strategy}. For example, we could define $G(\pi_0) := \frac{P(\pi = \pi_0 | URS)}{P(\pi = \pi_0 | UPS)}$. We use Bayes' rule to reduce this definition to $G(\pi_0) = \frac{P(URS | \pi = \pi_0)}{1 - P(URS | \pi = \pi_0)}$, as described in section \ref{Estimation classifier}, given a setting where $P(URS) = P(UPS)$. We then train a binary classifier to give a probability value for $P(URS | \pi = \pi_0)$. 

If the distributions of policies of these two sampling strategies are different (which we show later to be true in deterministic MDPs with self-loops), then policies with high goal-directedness will tend to have distinct internal ``features” of reward maximization that don’t show up randomly.


\section{Experiments and Results}

We now present three experiments of training a goal-directedness classifier under three stages of complexity: hand-picked MDP environments, a simple RL setting, and fine-tuned LLMs. We describe the first and third such settings here, and the second setting in section \ref{Miscellaneous RL results}. At each stage, we gradually make more assumptions and loosen more restrictions on our mathematical model as described in section \ref{Implications on experimental setup}, which makes computation easier at the risk of increasing the number of confounding variables. Throughout the rest of the paper, we use $\frac{P(USS)}{P(URS)}$ as a shorthand for the goal-directedness definition $G(\pi_0) := \frac{P(\pi = \pi_0 | USS)}{P(\pi = \pi_0 | URS)}$, and similarly for $\frac{P(URS)}{P(UPS)}$ and $\frac{P(USS)}{P(UPS)}$. 


Establishing ground truth for goal-directedness is challenging due to varying definitions. In this section, we provide provisional metrics and assess the classifier's performance through its loss and generalization across datasets. We find in our tests that these classifiers properly measure a consistent property that is goal-directedness and do not \textit{overfit} to confounding variables, especially when we approximate our classifier training process from our toy MDP definition. When the classifiers successfully generalize and identify goal-directed behaviors, it offers preliminary evidence that our approximation of the ideal classifier training process effectively captures goal-directedness. Future work would verify this method on more datasets, architectures, and training regimes.

\subsection{Markov Decision Process experiments}

We now train a classifier of our goal-directedness metric under randomly generated MDPs with certain structural properties. Specifically, consider a deterministic MDP, such that each transition $T(s, a, s’)$ has either probability $0$ or $1$, with guaranteed self-loops (i.e. for any $s$, there exists an action $a$ such that $T(s, a, s) = 1$). As a case study, let $|S| = 10$, $|A| = 4$, and $\gamma = 0.9$.

We use the Python MDP toolbox (\cite{MDPtoolbox}) to generate $10^4$ different MDPs and pick a $k \sim U[1, |T|]$, then $k$ rewards using URS. We then solve half of the MDPs to get half of our optimal policies, and randomize the other half, while labeling which were solved for and which were randomized. Then by default $P(USS | \pi = \pi_0) = P(UPS | \pi = \pi_0) = 0.5$. We use two classifier structures: a 3-layer, 64-width sequential neural network, and binary logistic regression. We then input certain features that intuitively seem relevant to the classifier, as defined in more detail in section \ref{MDP experiment terminology}.

\begin{figure}
  \centering
  \begin{subfigure}[b]{0.5\textwidth}
    \centering
    \includegraphics[width=\linewidth]{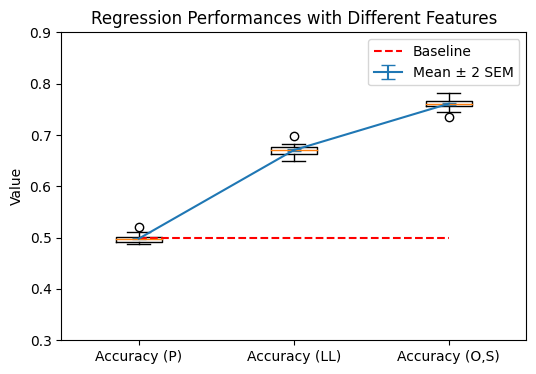}
    \caption{Accuracy graph}
    \label{fig:sub1}
  \end{subfigure}%
  \begin{subfigure}[b]{0.5\textwidth}
    \centering
    \includegraphics[width=\linewidth]{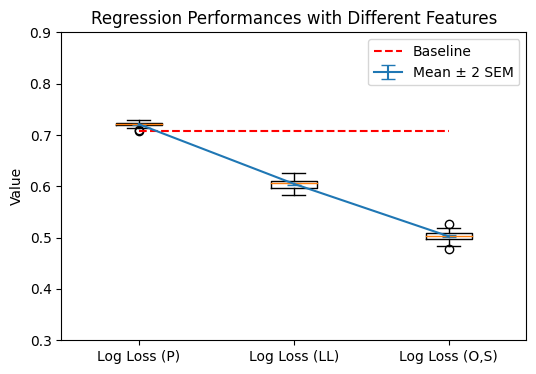}
    \caption{Loss graph}
    \label{fig:sub2}
  \end{subfigure}
  \caption{Accuracy and loss when passing in three different sets of features into the logistic classifier for predicting $\frac{P(\pi = \pi_0 | URS)}{P(\pi = \pi_0 | UPS)}$. Columns: (P) = policy (plus the flattened transition matrix and discount rate, although in practice it does not make a difference); (LL) = distance to loop and length of loop for the policy at each starting state $\pi(s_0)$; (O, S) = the sum of out-arrows, or states reachable from $s_0$, and whether $\pi(s_0)=s_0$ is a self-loop for any $s_0 \in S$. All error bars in the MDP experiments assume a normal distribution and show the two-sigma error of the independent generation of 30 classifiers for each category. All error bars in the MDP and RL experiments were calculated with calls to NumPy functions.}
  \label{fig:main1}
\end{figure}

Our results are shown in Figures \ref{fig:main1} and \ref{fig:main2}.\footnote{See figures \ref{fig:blownup1} and \ref{fig:blownup2} for bigger graphs.} For this task of determining whether a policy was generated via UPS or USS, we find that self-loops is the most predictive feature, followed by out-arrows visited, then distance to loop. These features correlate with features of power-seeking (POWER) and optimal policies found in \cite{turner2023optimal}, and indicate power-seeking and agency in toy settings. Policies that are rated as more goal-directed tend to reach more out-arrows and fewer self-loops, as predicted by \cite{turner2023optimal}. Additionally, combining features does not give a significant performance boost (appx. 0.01-0.04 accuracy boost by run). Finally, the neural network did not give a significant performance boost over the logistic classifier, suggesting that a different architecture is needed for better classifier performance.

In summary, we show that hand-crafted features that indicate ``power-seeking”, goal-directed behavior in a policy correlate with our metric when fed into the classifier. This shows that our metric is connected to properties like ``power-seeking" and optimality that we expect goal-directed agents to have. More findings are presented in section \ref{Miscellaneous MDP results}.

\subsection{LLM experiments}
\label{LLM experiments}

To investigate whether the mathematical definitions of goal-directedness extend to large neural networks, we conducted experiments using the open-source LLaMA-2-7B model developed by Meta (\cite{touvron2023llama2}). These experiments were designed to evaluate how well the concepts of sparse and dense reward functions, as described in the appendix (see Section \ref{Implications on experimental setup}), generalize when applied to LLMs. These loss functions are designed to be analogous to sparse versus dense reward functions without strictly adhering to the specific structure of the MDP framework. This corresponds to the distinction found in our $\frac{P(USS)}{P(URS)}$ metric. The exact definitions of these loss functions can be found in Section \ref{Loss Funcs}. 

We evaluated the performance of classifiers trained to predict our approximate $\frac{P(USS)}{P(URS)}$, or activations from models trained on dense versus sparse loss functions, on two distinct datasets, the GSM8K (\cite{cobbe2021gsm8k}) and Orca (\cite{mitra2024orcamath}) datasets, when applied to re-fine-tuned LLaMA-7B models (\cite{touvron2023llama}). To fine-tune these models, we use the LoRA method (\cite{hu2021lora}) on all LLMs. If our $\frac{P(USS)}{P(URS)}$ metric captures a real phenomenon and is not noise, then the classifier should be tractable to train and generalize to classify sparse loss function-trained models on other datasets.

We find that, although there is some noise in generalization, our goal-directedness classifier is able to separate sparse from dense loss function-trained models as having a higher probability of being sparsely trained across our tests, with a clear difference between (from high to low) the sparse 1-token, sparse 10-token, and dense function-trained models. In general, we give preliminary evidence that we can develop goal-directedness classifiers for LLMs that generalize between datasets. Implementation details and more findings are presented in section \ref{Miscellaneous LLM results}.

\begin{figure}[H]
    \centering
    \begin{subfigure}[b]{0.49\textwidth}
        \centering
        \includegraphics[width=\linewidth]{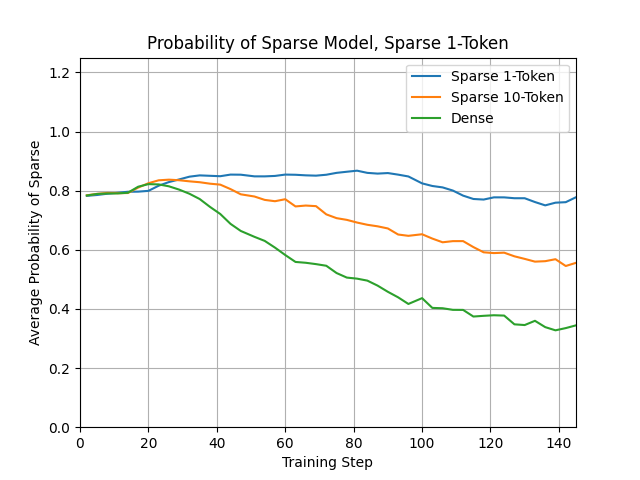}
        \label{fig:gsm8k_orca_one}
        \caption{Single token experiment.}
    \end{subfigure}
    \begin{subfigure}[b]{0.49\textwidth}
        \centering
        \includegraphics[width=\linewidth]{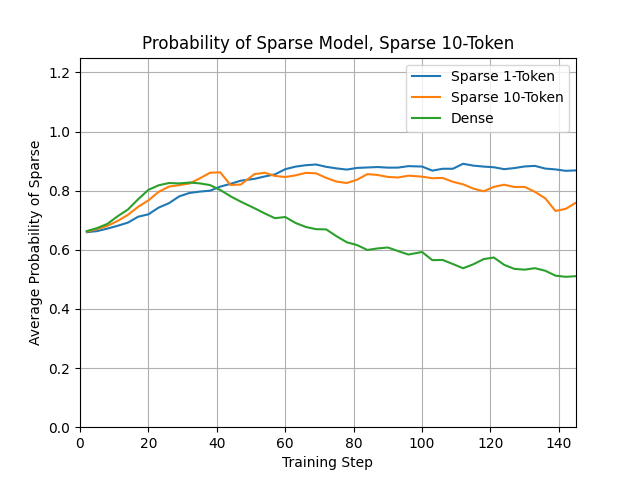}
        \label{fig:gsm8k_orca_ten}
        \caption{Ten token experiment.}
    \end{subfigure}
    \caption{Ablations for the LLM experiments. We train a classifier here on activations from two versions of llama-2-7b fine-tuned on GSM8k with sparse loss using one random token in the sequence and dense loss using all tokens respectively. We then tested the classifier on activations from models fine-tuned on a different dataset (Orca math) with sparse loss on one token, sparse loss on ten tokens, and dense loss. The classifier correctly separates these cases and places the denser ten-token-trained models between the 1-token and dense models. The right figure is generated similarly, except the sparsely-trained model creating data points for the classifier is trained on ten random tokens in the sequence. The classifier correctly separates these cases and identifies activations from models trained on one random token per sequence (which was not in the training distribution) as more sparse than ten-token-trained models. Rerunning the experiments with a different seed produces negligible variation in results.}
    \label{fig:gsm8K_orca}
\end{figure}

\section{Discussion}

In this work, we provide an intuitive definition of goal-directedness that is more tractable to compute than existing definitions, and a method to create a goal-directedness classifier. We first show that within an MDP setting, our definition is measurable and correlates with features associated with power-seeking. Features like a deterministic policy not taking self-loops and maximizing the number of out-arrows visited at each state tend to occur in optimal, power-seeking policies as shown by \cite{turner2023optimal}, and also occur in goal-directed policies according to our methods. We then adapt our methods to RL environments and again provide evidence that our definition is measurable in OpenAI Gym games (\cite{brockman2016openai}).

We also provide a demonstration of our method on LLMs, where we conducted experiments to measure the classifier's ability to identify the LoRA activations from different levels of sparse and dense loss functions, as defined in Section \ref{equation:dense}. Our preliminary results indicate that it is possible to achieve an accuracy greater than 70 percent for the dense loss function. As shown in Figure \ref{fig:gsm8K_orca}, when a model is fine-tuned on the loss signal from a specific dataset (GSM8K \cite{cobbe2021gsm8k} and the Orca Math dataset \cite{mitra2024orcamath}), and a classifier is trained on a different dataset, the classifier's ability to accurately detect signals within the activations of the sparse or dense loss function increases with the training steps.


We propose that goal-directedness is a requirement for advanced AIs pursuing unintended goals, which could lead to catasrophic risk. In this paper, we reviewed the literature for definitions of goal-directedness and created our own preliminary definition that fits the intentional stance and is agnostic about the agent's internal structure, while also being computationally efficient. We experiment on building classifiers based on our definition in increasingly complex environments and find preliminary evidence that these classifiers are tractable and properly generalize. We strongly recommend future research in this direction, and we provide more details in our recommendations in the appendix \ref{subsec:recommendations}. Our Social Impacts Statement is in section \ref{Broader impacts}. We further discuss conclusions in section \ref{Further discussions} and future work in section \ref{subsec:limitations}.

\subsection{Acknowledgements}

This project was created as part of the \href{https://www.constellation.org/programs/astra-fellowship}{Astra Fellowship} under the Winter 2024 Cohort, mentored by Richard Ngo. Our thanks go to Constellation for providing financial support and an office for much of this project. Thanks also to Martín Soto, Jeremy Gillen, Daniel Kokotajlo, Lukas Berglund, Gabriel Mukobi, Max Lamparth, and anonymous NeurIPS reviewers for feedback on various drafts of this paper.

\medskip

\bibliographystyle{plainnat}
\bibliography{references}

\begin{thebibliography}{78}
\providecommand{\natexlab}[1]{#1}
\providecommand{\url}[1]{\texttt{#1}}
\expandafter\ifx\csname urlstyle\endcsname\relax
  \providecommand{\doi}[1]{doi: #1}\else
  \providecommand{\doi}{doi: \begingroup \urlstyle{rm}\Url}\fi

\bibitem[MDP()]{MDPtoolbox}
Markov decision process (mdp) toolbox for python.
\newblock \url{https://pymdptoolbox.readthedocs.io/en/latest/index.html}.

\bibitem[Adam~Shimi(2021)]{litreview}
Joe~Collman Adam~Shimi, Michele~Campolo.
\newblock Literature review on goal-directedness.
\newblock \url{https://www.alignmentforum.org/posts/cfXwr6NC9AqZ9kr8g/literature-review-on-goal-directedness}, Jan 2021.
\newblock Accessed: 2024-05-09.

\bibitem[Allais(1990)]{Allais1990}
Maurice Allais.
\newblock \emph{Allais Paradox}, pages 3--9.
\newblock Palgrave Macmillan UK, London, 1990.
\newblock ISBN 978-1-349-20568-4.
\newblock \doi{10.1007/978-1-349-20568-4_2}.
\newblock URL \url{https://doi.org/10.1007/978-1-349-20568-4_2}.

\bibitem[Anthropic(2023)]{anthropicrsp}
Anthropic.
\newblock Anthropic's responsible scaling policy.
\newblock \url{https://www-cdn.anthropic.com/1adf000c8f675958c2ee23805d91aaade1cd4613/responsible-scaling-policy.pdf}, September 2023.

\bibitem[Barnett(2023)]{valuemisspecification}
Matthew Barnett.
\newblock Evaluating the historical value misspecification argument.
\newblock \url{https://www.alignmentforum.org/posts/i5kijcjFJD6bn7dwq/evaluating-the-historical-value-misspecification-argument}, Oct 2023.

\bibitem[BELLMAN and Dreyfus(2010)]{bellmaneq}
RICHARD BELLMAN and Stuart Dreyfus.
\newblock \emph{Dynamic Programming}, volume~33.
\newblock Princeton University Press, 2010.
\newblock ISBN 9780691146683.
\newblock URL \url{http://www.jstor.org/stable/j.ctv1nxcw0f}.

\bibitem[Belrose and Pope(2024)]{nocountingargs}
Nora Belrose and Quintin Pope.
\newblock Counting arguments provide no evidence for ai doom.
\newblock \url{https://optimists.ai/2024/02/27/counting-arguments-provide-no-evidence-for-ai-doom/}, Feb 2024.

\bibitem[Biehl and Virgo(2022)]{biehl2022interpretingsystemssolvingpomdps}
Martin Biehl and Nathaniel Virgo.
\newblock Interpreting systems as solving pomdps: a step towards a formal understanding of agency, 2022.
\newblock URL \url{https://arxiv.org/abs/2209.01619}.

\bibitem[Biewald(2020)]{wandb}
Lukas Biewald.
\newblock Experiment tracking with weights and biases, 2020.
\newblock URL \url{https://www.wandb.com/}.
\newblock Software available from wandb.com.

\bibitem[Bricken et~al.(2023)Bricken, Batson, Templeton, Jermyn, Henighan, and Olah]{featuresdef}
Trenton Bricken, Joshua Batson, Adly Templeton, Adam Jermyn, Tom Henighan, and Chris Olah.
\newblock Features as the simplest factorization.
\newblock \url{https://transformer-circuits.pub/2023/may-update/index.html#simple-factorization}, May 2023.

\bibitem[Brockman et~al.(2016)Brockman, Cheung, Pettersson, Schneider, Schulman, Tang, and Zaremba]{brockman2016openai}
Greg Brockman, Vicki Cheung, Ludwig Pettersson, Jonas Schneider, John Schulman, Jie Tang, and Wojciech Zaremba.
\newblock Openai gym, 2016.

\bibitem[Broome(1990)]{Bolker-Jeffrey}
John Broome.
\newblock {Bolker-Jeffrey Expected Utility Theory and Axiomatic Utilitarianism}.
\newblock \emph{The Review of Economic Studies}, 57\penalty0 (3):\penalty0 477--502, 07 1990.
\newblock ISSN 0034-6527.
\newblock \doi{10.2307/2298025}.
\newblock URL \url{https://doi.org/10.2307/2298025}.

\bibitem[Brown et~al.(2020)Brown, Mann, Ryder, Subbiah, Kaplan, Dhariwal, Neelakantan, Shyam, Sastry, Askell, Agarwal, Herbert-Voss, Krueger, Henighan, Child, Ramesh, Ziegler, Wu, Winter, Hesse, Chen, Sigler, Litwin, Gray, Chess, Clark, Berner, McCandlish, Radford, Sutskever, and Amodei]{brown2020language}
Tom~B. Brown, Benjamin Mann, Nick Ryder, Melanie Subbiah, Jared Kaplan, Prafulla Dhariwal, Arvind Neelakantan, Pranav Shyam, Girish Sastry, Amanda Askell, Sandhini Agarwal, Ariel Herbert-Voss, Gretchen Krueger, Tom Henighan, Rewon Child, Aditya Ramesh, Daniel~M. Ziegler, Jeffrey Wu, Clemens Winter, Christopher Hesse, Mark Chen, Eric Sigler, Mateusz Litwin, Scott Gray, Benjamin Chess, Jack Clark, Christopher Berner, Sam McCandlish, Alec Radford, Ilya Sutskever, and Dario Amodei.
\newblock Language models are few-shot learners, 2020.

\bibitem[Butlin(2022)]{Butlin2022MachineLF}
Patrick Butlin.
\newblock Machine learning, functions and goals.
\newblock \emph{Croatian journal of philosophy}, 2022.
\newblock URL \url{https://api.semanticscholar.org/CorpusID:255258483}.

\bibitem[Carlsmith(2023)]{carlsmith2023scheming}
Joe Carlsmith.
\newblock Scheming ais: Will ais fake alignment during training in order to get power?, 2023.

\bibitem[Choi and Kim(2015)]{ChoiBayesianIRL}
Jaedeug Choi and Kee-Eung Kim.
\newblock Hierarchical bayesian inverse reinforcement learning.
\newblock \emph{IEEE Transactions on Cybernetics}, 45\penalty0 (4):\penalty0 793--805, 2015.
\newblock \doi{10.1109/TCYB.2014.2336867}.

\bibitem[Cobbe et~al.(2021)Cobbe, Kosaraju, Bavarian, Chen, Jun, Kaiser, Plappert, Tworek, Hilton, Nakano, Hesse, and Schulman]{cobbe2021gsm8k}
Karl Cobbe, Vineet Kosaraju, Mohammad Bavarian, Mark Chen, Heewoo Jun, Lukasz Kaiser, Matthias Plappert, Jerry Tworek, Jacob Hilton, Reiichiro Nakano, Christopher Hesse, and John Schulman.
\newblock Training verifiers to solve math word problems.
\newblock \emph{arXiv preprint arXiv:2110.14168}, 2021.

\bibitem[da~Costa-Luis et~al.(2024)da~Costa-Luis, Larroque, Altendorf, Mary, richardsheridan, Korobov, Yorav-Raphael, Ivanov, Bargull, Rodrigues, Chen, Dektyarev, mjstevens777, Pagel, Zugnoni, JC, CrazyPython, Newey, Lee, pgajdos, Todd, Malmgren, redbug312, Desh, Nechaev, Górny, Boyle, Nordlund, MapleCCC, and McCracken]{casper_da_costa_luis_2024_11107065}
Casper da~Costa-Luis, Stephen~Karl Larroque, Kyle Altendorf, Hadrien Mary, richardsheridan, Mikhail Korobov, Noam Yorav-Raphael, Ivan Ivanov, Marcel Bargull, Nishant Rodrigues, Guangshuo Chen, Mikhail Dektyarev, mjstevens777, Matthew~D. Pagel, Martin Zugnoni, JC, CrazyPython, Charles Newey, Antony Lee, pgajdos, Todd, Staffan Malmgren, redbug312, Orivej Desh, Nikolay Nechaev, Michał Górny, Mike Boyle, Max Nordlund, MapleCCC, and Jack McCracken.
\newblock {tqdm: A fast, Extensible Progress Bar for Python and CLI}, May 2024.
\newblock URL \url{https://doi.org/10.5281/zenodo.11107065}.

\bibitem[de~Witt et~al.(2023)de~Witt, Sokota, Kolter, Foerster, and Strohmeier]{dewitt2023perfectly}
Christian~Schroeder de~Witt, Samuel Sokota, J.~Zico Kolter, Jakob Foerster, and Martin Strohmeier.
\newblock Perfectly secure steganography using minimum entropy coupling, 2023.

\bibitem[Deepmind()]{deepmindagi}
Deepmind.
\newblock \url{https://deepmind.google/about/}.

\bibitem[Dennett(2009)]{dennettintentional}
D.~Dennett.
\newblock Intentional systems theory.
\newblock \emph{The Oxford Handbook of Philosophy of Mind}, 01 2009.
\newblock \doi{10.1093/oxfordhb/9780199262618.003.0020}.

\bibitem[Dung(2024)]{Dung2024UnderstandingAA}
Le~Kim Dung.
\newblock Understanding artificial agency.
\newblock \emph{The Philosophical Quarterly}, 2024.
\newblock URL \url{https://api.semanticscholar.org/CorpusID:267564011}.

\bibitem[Farquhar et~al.(2022)Farquhar, Carey, and Everitt]{farquhar2022pathspecific}
Sebastian Farquhar, Ryan Carey, and Tom Everitt.
\newblock Path-specific objectives for safer agent incentives, 2022.

\bibitem[Goyal and Bengio(2022)]{inductivebiases}
Anirudh Goyal and Yoshua Bengio.
\newblock Inductive biases for deep learning of higher-level cognition.
\newblock \emph{Proceedings of the Royal Society A}, Oct 2022.

\bibitem[Hanheide et~al.(2010)Hanheide, Hawes, Wyatt, G{\"o}belbecker, Brenner, Sj{\"o}{\"o}, Aydemir, Jensfelt, Zender, and Kruijff]{Hanheide2010AFF}
Marc Hanheide, Nick Hawes, Jeremy~L. Wyatt, Moritz G{\"o}belbecker, Michael Brenner, Kristoffer Sj{\"o}{\"o}, Alper Aydemir, Patric Jensfelt, Hendrik Zender, and Geert-Jan~M. Kruijff.
\newblock A framework for goal generation and management.
\newblock In \emph{AAAI Conference on Artificial Intelligence}, 2010.
\newblock URL \url{https://api.semanticscholar.org/CorpusID:9741973}.

\bibitem[Harris et~al.(2020)Harris, Millman, van~der Walt, Gommers, Virtanen, Cournapeau, Wieser, Taylor, Berg, Smith, Kern, Picus, Hoyer, van Kerkwijk, Brett, Haldane, del Río, Wiebe, Peterson, Gérard-Marchant, Sheppard, Reddy, Weckesser, Abbasi, Gohlke, and Oliphant]{Harris_2020}
Charles~R. Harris, K.~Jarrod Millman, Stéfan~J. van~der Walt, Ralf Gommers, Pauli Virtanen, David Cournapeau, Eric Wieser, Julian Taylor, Sebastian Berg, Nathaniel~J. Smith, Robert Kern, Matti Picus, Stephan Hoyer, Marten~H. van Kerkwijk, Matthew Brett, Allan Haldane, Jaime~Fernández del Río, Mark Wiebe, Pearu Peterson, Pierre Gérard-Marchant, Kevin Sheppard, Tyler Reddy, Warren Weckesser, Hameer Abbasi, Christoph Gohlke, and Travis~E. Oliphant.
\newblock Array programming with numpy.
\newblock \emph{Nature}, 585\penalty0 (7825):\penalty0 357–362, September 2020.
\newblock ISSN 1476-4687.
\newblock \doi{10.1038/s41586-020-2649-2}.
\newblock URL \url{http://dx.doi.org/10.1038/s41586-020-2649-2}.

\bibitem[Heylighen(2022)]{Heylighen2022TheMA}
Francis Heylighen.
\newblock The meaning and origin of goal-directedness: a dynamical systems perspective.
\newblock \emph{Biological Journal of the Linnean Society}, 2022.
\newblock URL \url{https://api.semanticscholar.org/CorpusID:249695429}.

\bibitem[Hu et~al.(2021)Hu, Shen, Wallis, Allen-Zhu, Li, Wang, Wang, and Chen]{hu2021lora}
Edward~J. Hu, Yelong Shen, Phillip Wallis, Zeyuan Allen-Zhu, Yuanzhi Li, Shean Wang, Lu~Wang, and Weizhu Chen.
\newblock Lora: Low-rank adaptation of large language models, 2021.

\bibitem[Hubinger et~al.(2019)Hubinger, van Merwijk, Mikulik, Skalse, and Garrabrant]{hubinger2019risksfrom}
Evan Hubinger, Chris van Merwijk, Vladimir Mikulik, Joar Skalse, and Scott Garrabrant.
\newblock Risks from learned optimization in advanced machine learning systems, 2019.

\bibitem[Hubinger et~al.(2023)Hubinger, Schiefer, Denison, and Perez]{hubinger2023model}
Evan Hubinger, Nicholas Schiefer, Carson Denison, and Ethan Perez.
\newblock Model organisms of misalignment: The case for a new pillar of alignment research.
\newblock \url{https://www.alignmentforum.org/posts/ChDH335ckdvpxXaXX/model-organisms-of-misalignment-the-case-for-a-new-pillar-of-1}, Aug 2023.
\newblock Accessed: 2024-05-04.

\bibitem[Hunter(2007)]{Hunter:2007}
J.~D. Hunter.
\newblock Matplotlib: A 2d graphics environment.
\newblock \emph{Computing in Science \& Engineering}, 9\penalty0 (3):\penalty0 90--95, 2007.
\newblock \doi{10.1109/MCSE.2007.55}.

\bibitem[Hutter(2000)]{hutter2000theory}
Marcus Hutter.
\newblock A theory of universal artificial intelligence based on algorithmic complexity, 2000.

\bibitem[Jumper et~al.(2021)Jumper, Evans, Pritzel, et~al.]{JumperAlphafold2021}
John Jumper, Richard Evans, Alexander Pritzel, et~al.
\newblock Highly accurate protein structure prediction with alphafold.
\newblock \emph{Nature}, 596:\penalty0 583--589, 2021.
\newblock \doi{10.1038/s41586-021-03819-2}.
\newblock URL \url{https://doi.org/10.1038/s41586-021-03819-2}.

\bibitem[Kenton et~al.(2023)Kenton, Kumar, Farquhar, Richens, MacDermott, and Everitt]{KENTON2023103963}
Zachary Kenton, Ramana Kumar, Sebastian Farquhar, Jonathan Richens, Matt MacDermott, and Tom Everitt.
\newblock Discovering agents.
\newblock \emph{Artificial Intelligence}, 322:\penalty0 103963, 2023.
\newblock ISSN 0004-3702.
\newblock \doi{https://doi.org/10.1016/j.artint.2023.103963}.
\newblock URL \url{https://www.sciencedirect.com/science/article/pii/S0004370223001091}.

\bibitem[Kolmogorov(1998)]{Kolmogorov1998OnTO}
Andrei~N. Kolmogorov.
\newblock On tables of random numbers (reprinted from "sankhya: The indian journal of statistics", series a, vol. 25 part 4, 1963).
\newblock \emph{Theor. Comput. Sci.}, 207:\penalty0 387--395, 1998.
\newblock URL \url{https://api.semanticscholar.org/CorpusID:33390800}.

\bibitem[Kosoy(2023)]{kosoydefinition}
Vanessa Kosoy.
\newblock The learning-theoretic agenda: Status 2023.
\newblock \url{https://www.alignmentforum.org/posts/ZwshvqiqCvXPsZEct/the-learning-theoretic-agenda-status-2023%23Direction_17__Algorithmic_Descriptive_Agency_Measure__ADAM_}, Apr 2023.
\newblock Accessed: 2024-05-09.

\bibitem[Langosco et~al.(2023)Langosco, Koch, Sharkey, Pfau, Orseau, and Krueger]{langosco2023goal}
Lauro Langosco, Jack Koch, Lee Sharkey, Jacob Pfau, Laurent Orseau, and David Krueger.
\newblock Goal misgeneralization in deep reinforcement learning, 2023.

\bibitem[Lhoest et~al.(2021)Lhoest, Villanova~del Moral, Jernite, Thakur, von Platen, Patil, Chaumond, Drame, Plu, Tunstall, Davison, {\v{S}}a{\v{s}}ko, Chhablani, Malik, Brandeis, Le~Scao, Sanh, Xu, Patry, McMillan-Major, Schmid, Gugger, Delangue, Matussi{\`e}re, Debut, Bekman, Cistac, Goehringer, Mustar, Lagunas, Rush, and Wolf]{lhoest-etal-2021-datasets}
Quentin Lhoest, Albert Villanova~del Moral, Yacine Jernite, Abhishek Thakur, Patrick von Platen, Suraj Patil, Julien Chaumond, Mariama Drame, Julien Plu, Lewis Tunstall, Joe Davison, Mario {\v{S}}a{\v{s}}ko, Gunjan Chhablani, Bhavitvya Malik, Simon Brandeis, Teven Le~Scao, Victor Sanh, Canwen Xu, Nicolas Patry, Angelina McMillan-Major, Philipp Schmid, Sylvain Gugger, Cl{\'e}ment Delangue, Th{\'e}o Matussi{\`e}re, Lysandre Debut, Stas Bekman, Pierric Cistac, Thibault Goehringer, Victor Mustar, Fran{\c{c}}ois Lagunas, Alexander Rush, and Thomas Wolf.
\newblock Datasets: A community library for natural language processing.
\newblock In \emph{Proceedings of the 2021 Conference on Empirical Methods in Natural Language Processing: System Demonstrations}, pages 175--184, Online and Punta Cana, Dominican Republic, November 2021. Association for Computational Linguistics.
\newblock URL \url{https://aclanthology.org/2021.emnlp-demo.21}.

\bibitem[Luck and d'Inverno(1995)]{Luck1995AFF}
Michael Luck and Mark d'Inverno.
\newblock A formal framework for agency and autonomy.
\newblock In \emph{International Conference on Multiagent Systems}, 1995.
\newblock URL \url{https://api.semanticscholar.org/CorpusID:5357752}.

\bibitem[Mangrulkar et~al.(2022)Mangrulkar, Gugger, Debut, Belkada, Paul, and Bossan]{peft}
Sourab Mangrulkar, Sylvain Gugger, Lysandre Debut, Younes Belkada, Sayak Paul, and Benjamin Bossan.
\newblock Peft: State-of-the-art parameter-efficient fine-tuning methods.
\newblock \url{https://github.com/huggingface/peft}, 2022.

\bibitem[Mitra et~al.(2024)Mitra, Khanpour, Rosset, and Awadallah]{mitra2024orcamath}
Arindam Mitra, Hamed Khanpour, Corby Rosset, and Ahmed Awadallah.
\newblock Orca-math: Unlocking the potential of slms in grade school math, 2024.

\bibitem[Nanda et~al.(2023)Nanda, Lee, and Wattenberg]{nanda2023emergent}
Neel Nanda, Andrew Lee, and Martin Wattenberg.
\newblock Emergent linear representations in world models of self-supervised sequence models, 2023.

\bibitem[Ng and Russell(2000)]{IRLoriginal}
Andrew Ng and Stuart Russell.
\newblock Algorithms for inverse reinforcement learning.
\newblock \emph{ICML '00 Proceedings of the Seventeenth International Conference on Machine Learning}, 05 2000.

\bibitem[Ngo(2020)]{agisafetyprinciples}
Richard Ngo.
\newblock Agi safety from first principles.
\newblock \url{https://drive.google.com/file/d/1uK7NhdSKprQKZnRjU58X7NLA1auXlWHt/view}, Sep 2020.

\bibitem[Ngo et~al.(2022)Ngo, Chan, and Mindermann]{ngo2022deeplearning}
Richard Ngo, Lawrence Chan, and Soren Mindermann.
\newblock The alignment problem from a deep learning perspective, 2022.

\bibitem[Oesterheld(2015)]{Oesterheld_2015}
Caspar Oesterheld.
\newblock Formalizing preference utilitarianism in physical world models.
\newblock \emph{Synthese}, 193\penalty0 (9):\penalty0 2747–2759, September 2015.
\newblock ISSN 1573-0964.
\newblock \doi{10.1007/s11229-015-0883-1}.
\newblock URL \url{http://dx.doi.org/10.1007/s11229-015-0883-1}.

\bibitem[OpenAI()]{openaiagi}
OpenAI.
\newblock \url{https://openai.com/about/}.

\bibitem[Orseau et~al.(2018)Orseau, McGill, and Legg]{agentsanddevices}
Laurent Orseau, Simon~McGregor McGill, and Shane Legg.
\newblock Agents and devices: A relative definition of agency, 2018.

\bibitem[Paszke et~al.(2019)Paszke, Gross, Massa, Lerer, Bradbury, Chanan, Killeen, Lin, Gimelshein, Antiga, Desmaison, Köpf, Yang, DeVito, Raison, Tejani, Chilamkurthy, Steiner, Fang, Bai, and Chintala]{paszke2019pytorch}
Adam Paszke, Sam Gross, Francisco Massa, Adam Lerer, James Bradbury, Gregory Chanan, Trevor Killeen, Zeming Lin, Natalia Gimelshein, Luca Antiga, Alban Desmaison, Andreas Köpf, Edward Yang, Zach DeVito, Martin Raison, Alykhan Tejani, Sasank Chilamkurthy, Benoit Steiner, Lu~Fang, Junjie Bai, and Soumith Chintala.
\newblock Pytorch: An imperative style, high-performance deep learning library, 2019.

\bibitem[Pedregosa et~al.(2018)Pedregosa, Varoquaux, Gramfort, Michel, Thirion, Grisel, Blondel, Müller, Nothman, Louppe, Prettenhofer, Weiss, Dubourg, Vanderplas, Passos, Cournapeau, Brucher, Perrot, and Édouard Duchesnay]{pedregosa2018scikitlearn}
Fabian Pedregosa, Gaël Varoquaux, Alexandre Gramfort, Vincent Michel, Bertrand Thirion, Olivier Grisel, Mathieu Blondel, Andreas Müller, Joel Nothman, Gilles Louppe, Peter Prettenhofer, Ron Weiss, Vincent Dubourg, Jake Vanderplas, Alexandre Passos, David Cournapeau, Matthieu Brucher, Matthieu Perrot, and Édouard Duchesnay.
\newblock Scikit-learn: Machine learning in python, 2018.

\bibitem[Puterman(1994)]{MDPintro}
Martin~L. Puterman.
\newblock \emph{Markov Decision Processes: Discrete Stochastic Dynamic Programming}.
\newblock John Wiley I\& Sons, Apr 1994.
\newblock \doi{10.1002/9780470316887}.

\bibitem[Rafailov et~al.(2024)Rafailov, Sharma, Mitchell, Ermon, Manning, and Finn]{rafailov2024dpo}
Rafael Rafailov, Archit Sharma, Eric Mitchell, Stefano Ermon, Christopher~D. Manning, and Chelsea Finn.
\newblock Direct preference optimization: Your language model is secretly a reward model, 2024.
\newblock URL \url{https://arxiv.org/abs/2305.18290}.

\bibitem[Rahaman et~al.(2019)Rahaman, Baratin, Arpit, Draxler, Lin, Hamprecht, Bengio, and Courville]{spectralbias}
Nasim Rahaman, Aristide Baratin, Devansh Arpit, Felix Draxler, Min Lin, Fred Hamprecht, Yoshua Bengio, and Aaron Courville.
\newblock On the spectral bias of neural networks.
\newblock In Kamalika Chaudhuri and Ruslan Salakhutdinov, editors, \emph{Proceedings of the 36th International Conference on Machine Learning}, volume~97 of \emph{Proceedings of Machine Learning Research}, pages 5301--5310. PMLR, 09--15 Jun 2019.
\newblock URL \url{https://proceedings.mlr.press/v97/rahaman19a.html}.

\bibitem[Ramesh et~al.(2021)Ramesh, Pavlov, Goh, Gray, Voss, Radford, Chen, and Sutskever]{ramesh2021zeroshot}
Aditya Ramesh, Mikhail Pavlov, Gabriel Goh, Scott Gray, Chelsea Voss, Alec Radford, Mark Chen, and Ilya Sutskever.
\newblock Zero-shot text-to-image generation, 2021.

\bibitem[Robert(2007)]{EDTintro}
Christian Robert.
\newblock \emph{The Bayesian Choice: From Decision Theoretic Foundations to Computational Implementation}.
\newblock Springer, 01 2007.

\bibitem[Savage(1954)]{savage54}
Leonard Savage.
\newblock \emph{The Foundations of Statistics}.
\newblock Wiley, 1954.

\bibitem[Shah et~al.(2022)Shah, Varma, Kumar, Phuong, Krakovna, Uesato, and Kenton]{shah2022goal}
Rohin Shah, Vikrant Varma, Ramana Kumar, Mary Phuong, Victoria Krakovna, Jonathan Uesato, and Zac Kenton.
\newblock Goal misgeneralization: Why correct specifications aren't enough for correct goals, 2022.

\bibitem[Silver et~al.(2016)Silver, Huang, Maddison, et~al.]{SilverGo2016}
David Silver, Aja Huang, Chris Maddison, et~al.
\newblock Mastering the game of go with deep neural networks and tree search.
\newblock \emph{Nature}, 529:\penalty0 484--489, 2016.
\newblock \doi{10.1038/nature16961}.
\newblock URL \url{https://doi.org/10.1038/nature16961}.

\bibitem[Solomonoff(1964)]{SOLOMONOFF19641}
R.J. Solomonoff.
\newblock A formal theory of inductive inference. part i.
\newblock \emph{Information and Control}, 7\penalty0 (1):\penalty0 1--22, 1964.
\newblock ISSN 0019-9958.
\newblock \doi{https://doi.org/10.1016/S0019-9958(64)90223-2}.
\newblock URL \url{https://www.sciencedirect.com/science/article/pii/S0019995864902232}.

\bibitem[Touvron et~al.(2023{\natexlab{a}})Touvron, Lavril, Izacard, Martinet, Lachaux, Lacroix, Rozière, Goyal, Hambro, Azhar, Rodriguez, Joulin, Grave, and Lample]{touvron2023llama}
Hugo Touvron, Thibaut Lavril, Gautier Izacard, Xavier Martinet, Marie-Anne Lachaux, Timothée Lacroix, Baptiste Rozière, Naman Goyal, Eric Hambro, Faisal Azhar, Aurelien Rodriguez, Armand Joulin, Edouard Grave, and Guillaume Lample.
\newblock Llama: Open and efficient foundation language models, 2023{\natexlab{a}}.

\bibitem[Touvron et~al.(2023{\natexlab{b}})Touvron, Martin, Stone, Albert, Almahairi, Babaei, Bashlykov, Batra, Bhargava, Bhosale, Bikel, Blecher, Ferrer, Chen, Cucurull, Esiobu, Fernandes, Fu, Fu, Fuller, Gao, Goswami, Goyal, Hartshorn, Hosseini, Hou, Inan, Kardas, Kerkez, Khabsa, Kloumann, Korenev, Koura, Lachaux, Lavril, Lee, Liskovich, Lu, Mao, Martinet, Mihaylov, Mishra, Molybog, Nie, Poulton, Reizenstein, Rungta, Saladi, Schelten, Silva, Smith, Subramanian, Tan, Tang, Taylor, Williams, Kuan, Xu, Yan, Zarov, Zhang, Fan, Kambadur, Narang, Rodriguez, Stojnic, Edunov, and Scialom]{touvron2023llama2}
Hugo Touvron, Louis Martin, Kevin Stone, Peter Albert, Amjad Almahairi, Yasmine Babaei, Nikolay Bashlykov, Soumya Batra, Prajjwal Bhargava, Shruti Bhosale, Dan Bikel, Lukas Blecher, Cristian~Canton Ferrer, Moya Chen, Guillem Cucurull, David Esiobu, Jude Fernandes, Jeremy Fu, Wenyin Fu, Brian Fuller, Cynthia Gao, Vedanuj Goswami, Naman Goyal, Anthony Hartshorn, Saghar Hosseini, Rui Hou, Hakan Inan, Marcin Kardas, Viktor Kerkez, Madian Khabsa, Isabel Kloumann, Artem Korenev, Punit~Singh Koura, Marie-Anne Lachaux, Thibaut Lavril, Jenya Lee, Diana Liskovich, Yinghai Lu, Yuning Mao, Xavier Martinet, Todor Mihaylov, Pushkar Mishra, Igor Molybog, Yixin Nie, Andrew Poulton, Jeremy Reizenstein, Rashi Rungta, Kalyan Saladi, Alan Schelten, Ruan Silva, Eric~Michael Smith, Ranjan Subramanian, Xiaoqing~Ellen Tan, Binh Tang, Ross Taylor, Adina Williams, Jian~Xiang Kuan, Puxin Xu, Zheng Yan, Iliyan Zarov, Yuchen Zhang, Angela Fan, Melanie Kambadur, Sharan Narang, Aurelien Rodriguez, Robert Stojnic, Sergey Edunov, and Thomas
  Scialom.
\newblock Llama 2: Open foundation and fine-tuned chat models, 2023{\natexlab{b}}.

\bibitem[Turner and Tadepalli(2022)]{turner2022parametrically}
Alexander~Matt Turner and Prasad Tadepalli.
\newblock Parametrically retargetable decision-makers tend to seek power, 2022.

\bibitem[Turner et~al.(2023{\natexlab{a}})Turner, Smith, Shah, Critch, and Tadepalli]{turner2023optimal}
Alexander~Matt Turner, Logan Smith, Rohin Shah, Andrew Critch, and Prasad Tadepalli.
\newblock Optimal policies tend to seek power, 2023{\natexlab{a}}.

\bibitem[Turner et~al.(2023{\natexlab{b}})Turner, Thiergart, Udell, Leech, Mini, and MacDiarmid]{turner2023activation}
Alexander~Matt Turner, Lisa Thiergart, David Udell, Gavin Leech, Ulisse Mini, and Monte MacDiarmid.
\newblock Activation addition: Steering language models without optimization, 2023{\natexlab{b}}.

\bibitem[Udell(2022)]{udellshard}
David Udell.
\newblock Shard theory: An overview.
\newblock \url{https://www.alignmentforum.org/posts/xqkGmfikqapbJ2YMj/shard-theory-an-overview}, Aug 2022.
\newblock Accessed: 2024-05-09.

\bibitem[Van~Rossum(2020)]{van1995python}
Guido Van~Rossum.
\newblock \emph{The Python Library Reference, release 3.8.2}.
\newblock Python Software Foundation, 2020.

\bibitem[Van~Rossum and Drake(2009)]{10.5555/1593511}
Guido Van~Rossum and Fred~L. Drake.
\newblock \emph{Python 3 Reference Manual}.
\newblock CreateSpace, Scotts Valley, CA, 2009.
\newblock ISBN 1441412697.

\bibitem[Vaswani et~al.(2023)Vaswani, Shazeer, Parmar, Uszkoreit, Jones, Gomez, Kaiser, and Polosukhin]{vaswani2023attention}
Ashish Vaswani, Noam Shazeer, Niki Parmar, Jakob Uszkoreit, Llion Jones, Aidan~N. Gomez, Lukasz Kaiser, and Illia Polosukhin.
\newblock Attention is all you need, 2023.

\bibitem[Veit et~al.(2016)Veit, Wilber, and Belongie]{veit2016residual}
Andreas Veit, Michael Wilber, and Serge Belongie.
\newblock Residual networks behave like ensembles of relatively shallow networks, 2016.

\bibitem[Vinyals et~al.(2019)Vinyals, Babuschkin, Czarnecki, et~al.]{Vinyals2019}
Oriol Vinyals, Igor Babuschkin, Wojciech~M. Czarnecki, et~al.
\newblock Grandmaster level in starcraft ii using multi-agent reinforcement learning.
\newblock \emph{Nature}, 575:\penalty0 350--354, 2019.
\newblock \doi{10.1038/s41586-019-1724-z}.
\newblock URL \url{https://doi.org/10.1038/s41586-019-1724-z}.

\bibitem[von Neumann et~al.(1944)von Neumann, Morgenstern, and Rubinstein]{vnmtheorem}
John von Neumann, Oskar Morgenstern, and Ariel Rubinstein.
\newblock \emph{Theory of Games and Economic Behavior (60th Anniversary Commemorative Edition)}.
\newblock Princeton University Press, 1944.
\newblock ISBN 9780691130613.
\newblock URL \url{http://www.jstor.org/stable/j.ctt1r2gkx}.

\bibitem[Watkins(1989)]{qlearning}
Christopher Watkins.
\newblock Learning from delayed rewards.
\newblock 01 1989.

\bibitem[Wentworth(2024)]{wentworthcoherence}
John Wentworth.
\newblock Coherence of caches and agents.
\newblock \url{https://www.lesswrong.com/posts/wjFijaAkSCceqCgGF/coherence-of-caches-and-agents}, April 2024.

\bibitem[Wolf et~al.(2020)Wolf, Debut, Sanh, Chaumond, Delangue, Moi, Cistac, Rault, Louf, Funtowicz, Davison, Shleifer, von Platen, Ma, Jernite, Plu, Xu, Scao, Gugger, Drame, Lhoest, and Rush]{wolf-etal-2020-transformers}
Thomas Wolf, Lysandre Debut, Victor Sanh, Julien Chaumond, Clement Delangue, Anthony Moi, Pierric Cistac, Tim Rault, Rémi Louf, Morgan Funtowicz, Joe Davison, Sam Shleifer, Patrick von Platen, Clara Ma, Yacine Jernite, Julien Plu, Canwen Xu, Teven~Le Scao, Sylvain Gugger, Mariama Drame, Quentin Lhoest, and Alexander~M. Rush.
\newblock Transformers: State-of-the-art natural language processing.
\newblock In \emph{Proceedings of the 2020 Conference on Empirical Methods in Natural Language Processing: System Demonstrations}, pages 38--45, Online, October 2020. Association for Computational Linguistics.
\newblock URL \url{https://www.aclweb.org/anthology/2020.emnlp-demos.6}.

\bibitem[Yudkowsky(2017)]{yudkowsky2017coherent}
Eliezer Yudkowsky.
\newblock Coherent decisions imply consistent utilities.
\newblock \url{https://www.lesswrong.com/posts/RQpNHSiWaXTvDxt6R/coherent-decisions-imply-consistent-utilities}, 2017.
\newblock Originally written for Arbital. Accessed: 2024-05-04.

\bibitem[Zhou et~al.(2021)Zhou, Cui, Hu, Zhang, Yang, Liu, Wang, Li, and Sun]{zhou2021graph}
Jie Zhou, Ganqu Cui, Shengding Hu, Zhengyan Zhang, Cheng Yang, Zhiyuan Liu, Lifeng Wang, Changcheng Li, and Maosong Sun.
\newblock Graph neural networks: A review of methods and applications, 2021.

\bibitem[Zou et~al.(2023)Zou, Phan, Chen, Campbell, Guo, Ren, Pan, Yin, Mazeika, Dombrowski, Goel, Li, Byun, Wang, Mallen, Basart, Koyejo, Song, Fredrikson, Kolter, and Hendrycks]{zou2023representation}
Andy Zou, Long Phan, Sarah Chen, James Campbell, Phillip Guo, Richard Ren, Alexander Pan, Xuwang Yin, Mantas Mazeika, Ann-Kathrin Dombrowski, Shashwat Goel, Nathaniel Li, Michael~J. Byun, Zifan Wang, Alex Mallen, Steven Basart, Sanmi Koyejo, Dawn Song, Matt Fredrikson, J.~Zico Kolter, and Dan Hendrycks.
\newblock Representation engineering: A top-down approach to ai transparency, 2023.

\bibitem[Świechowski et~al.(2022)Świechowski, Godlewski, Sawicki, and Mańdziuk]{mcts_wiechowski_2022}
Maciej Świechowski, Konrad Godlewski, Bartosz Sawicki, and Jacek Mańdziuk.
\newblock Monte carlo tree search: a review of recent modifications and applications.
\newblock \emph{Artificial Intelligence Review}, 56\penalty0 (3):\penalty0 2497–2562, July 2022.
\newblock ISSN 1573-7462.
\newblock \doi{10.1007/s10462-022-10228-y}.
\newblock URL \url{http://dx.doi.org/10.1007/s10462-022-10228-y}.

\end{thebibliography}

\newpage
\appendix
\section{Appendix}

\subsection{Social Impacts Statement}
\label{Broader impacts}
In this paper, we discuss goal-directedness to better understand and prevent worst-case risks from advanced AI systems. Understanding whether AIs are goal-directed is crucial to evaluating models and possibly preventing misalignment risks like goal misgeneralization. This work could also elucidate which misalignment threat models depending on goal-directedness are more or less likely. While this work could hypothetically enable future AI developers to build more goal-directed models, increasing worst-case risks, we believe the benefits of understanding and measuring goal-directedness outweigh the possible downsides.


\subsection{Implications on experimental setup}
\label{Implications on experimental setup}

We use the model as described above for our MDP experiments in Section 4.1, sampling optimal policies for each strategy. However, in order to apply our model to more complicated settings listed in section \ref{LLM experiments}, we adapt our model by changing some of the sampling methods. For instance, instead of using $\epsilon$-greedy policies, we ``sample" policies trained on their respective reward functions for near-optimal policy sampling. We also lean on a more general notion of what it means to have ``sparse" or ``dense" reward in section \ref{LLM experiments}. As such, we consider this model to give \textbf{a general guideline} for how to measure goal-directedness, rather than a strict formula. The key elements are \textit{optimality} and \textit{generality}, as measured by comparing URS to UPS, and \textit{sparsity} of the reward function, as measured by comparing USS to URS.

\subsection{Our mathematical model in more detail}
\label{Mathematical model}

For our mathematical definition, we work within the MDP (Markov Decision Process) framework (\cite{MDPintro}) with a set of states $S$, a set of possible actions $A$ that can be selected in each state, and a transition function $T(s, a)$ that returns a probability distribution over all states $s’ \in S$ (such that $T(s, a, s’) \in \mathbb{R}$), but \textit{without} a predefined reward function. Then, we can define a distribution from which we sample a reward function $R ~ D$, and since $R$ and the MDP are invariant across time-steps, we can define a (deterministic) policy $\pi \in [1, |A|]^{|S|}$ as a tuple of actions, one action to take for each state.

In this section, we will talk about deterministic MDPs and policies (as an example of environments and AIs, respectively). Whenever we talk about sampling from the space of policies, we will assume that this just samples uniformly from all combinations of discrete actions; we will call this \textbf{ uniform policy sampling (UPS)}. We sample rewards from some prior distribution $\zeta_r$; since optimal policies are invariant under scaling reward functions, we let $\zeta_r = U[-1,1]$ as an example. Thus let $D_{\zeta_r}=D_{U[-1,1]-\text{IID}}$ be the distribution of reward functions where each reward of each transition $R(s, a, s’)$ is drawn uniformly and independently from $U[-1, 1]$. We call this \textbf{Uniform Reward Sampling (URS)}. Even under URS, some policies will be more coherent than others, because they will be optimal for more reward functions.

\begin{definition}
For a given policy $\pi_0$ sampled from $\pi \sim URS$, we measure goal-directedness as follows (where $|\pi|$ is the number of possible policies):

\begin{equation}
G(\pi_0) := \frac{P(\pi = \pi_0 | URS)}{P(\pi = \pi_0 | UPS)} = P(\pi = \pi_0 | URS) |\pi|
\end{equation}
\end{definition}

This is a difficult function to estimate directly because enumerating over all computable reward functions is intractable, and small epsilon perturbations in a reward function can cause the optimal policy to change significantly. \footnote{For example, consider a policy that starts at a state $A$ and can get high reward by going to state $C$ through $B$ (such that going $A \to B \to C$ is optimal), but is indifferent between two paths from $A$ to $B$. Then an $\epsilon-$change in the rewards on one of the paths from $A$ to $B$ will rule out half of the optimal policies.} Instead, we can use the following indirect estimation technique.

\subsubsection{Estimation classifier}
\label{Estimation classifier}

In order to estimate $P(\pi = \pi_0 | URS)$, we first estimate the reverse. Specifically, consider a setting where we first flip a coin, then sample $\pi$ using URS if it is heads, and UPS if it is tails. In this setting, we can train a binary classifier $P(URS | \pi = \pi_0)$ by generating two sets of URS and UPS policies and using these sets as training data. But by Bayes’ theorem:

\begin{equation}
P(URS | \pi = \pi_0) = \frac{P(\pi=\pi_0 | URS)P(URS)}{P(\pi=\pi_0)} = \frac{0.5P(\pi=\pi_0 | URS)}{0.5P(\pi=\pi_0 | URS)+0.5P(\pi=\pi_0 | UPS)}
\end{equation}

\begin{equation}
P(URS | \pi = \pi_0) = \frac{P(\pi=\pi_0 | URS)}{P(\pi=\pi_0 | URS)+P(\pi=\pi_0 | UPS)}
\end{equation}

Rearranging gives:

\begin{equation}
    P(\pi = \pi_0 | URS) = P(URS | \pi = \pi_0)P(\pi = \pi_0 | URS) 
    + P(URS | \pi = \pi_0)P(\pi = \pi_0 | UPS)
\end{equation}

And so: $P(\pi = \pi_0 | URS)(1 - P(URS | \pi = \pi_0)) = P(URS | \pi = \pi_0)P(\pi = \pi_0 | UPS)$

Therefore, $G(\pi_0) = \frac{P(\pi=\pi_0 | URS)}{P(\pi=\pi_0 | UPS)} = \frac{P(URS|\pi=\pi_0)}{1-P(URS|\pi=\pi_0)}$.

In theory, the classifier learns to distinguish between more goal-directed and less goal-directed policies, leading to a goal-directedness classifier. Note that the correct classification of a policy may depend on the graph structure of the underlying MDP, in a way which is hard to capture with standard classifiers. Classifier architectures could include a graph neural network (\cite{zhou2021graph}), a simple linear/logistic classifier using hand-crafted features, and/or with interpretability tools like linear probes in the case of an underlying neural network as the policy network. 

\subsubsection{Sparsity as simplicity}

Intuitively speaking, we would still like to differentiate policies that are optimal for ``simple" reward functions, to avoid convoluted, degenerate reward functions explaining a policy's behavior. In the context of MDPs, if we only need to specify rewards for a few states or transitions, that is much simpler than specifying rewards for every state or transition. Thus our version of ``simplicity" in our model is the \textit{sparsity} of the reward function. 

Specifically, consider a new distribution over rewards: \textbf{Uniform Sparsity Sampling (USS)}. As a simplification of some simplicity prior $\zeta_s$, for an MDP with $N$ state-action transitions, we set some value $k$, either via $k \sim U[1,N]$ for section 4.1 or presetting $k = 0.01 \cdot N$ for section 4.2. We then sample random rewards for $k$ transitions (selected uniformly without replacement), and finally sample a policy which is optimal for that reward. All the equations work the same, meaning that:

\begin{equation}
G(\pi_0) = \frac{P(\pi=\pi_0 | USS)}{P(\pi=\pi_0 | UPS)} = \frac{P(USS|\pi=\pi_0)}{1-P(USS|\pi=\pi_0)}
\end{equation}

We can also define goal-directedness in this setting as $\frac{P(\pi = \pi_0 | USS)}{P(\pi = \pi_0 | URS)}$, which intuitively represents how sparse the reward functions that a policy is optimal for is given that it is already ``coherent” to some degree. (This is also our latter definition of goal-directedness divided by our former definition.) Doing the same calculation as in Section \ref{Estimation classifier} also gives us $G(\pi_0) = \frac{P(USS | \pi = \pi_0)}{1 - P(USS | \pi = \pi_0)}$, except that the choice is between USS and URS instead of UPS. Another way of generating sparse policies is by sampling rewards from a high-variance distribution, and possibly discarding the ones which are below a given threshold.

Under our setup, $G(\pi_0)$ ranges from 0 when $P(URS | \pi = \pi_0) = 0$ to $+\infty$ when $P(URS | \pi = \pi_0) = 1$; the prior, not knowing anything specific about $\pi_0$, is $P(URS | \pi = \pi_0) = 0.5$, implying $G(\pi_0) = 1$. Policies that are optimal, or almost optimal, for a broader class of reward functions will have higher $P(\pi = \pi_0 | URS)$ and thus higher goal-directedness.

\subsubsection{Suboptimality}

The current method only counts a policy if it is \textit{exactly} optimal for a given reward function. But real-world agents will never be actually optimal for any non-trivial reward function. So if a policy is \textit{almost} optimal for many reward functions, that should still count towards its goal-directedness.

We can therefore add another step. Instead of only sampling from optimal policies for a given reward function, we could first sample a value $\epsilon \in (0, 1)$, then sample a policy which has expected reward within $\epsilon$ of the expected reward of the optimal policy (e.g. by early stopping). Note that this can be combined with different possibilities for how to do simplicity-weighted reward sampling. 

To recap, the methodology in this toy setting is:
\begin{enumerate}
    \item Sample a value $k$ which determines the simplicity of the reward function, or what percent of $R(s)$ or $R(s, a, s') \neq 0$, or otherwise have some simplicity prior $\zeta_s$ over the distribution of reward functions in the environment.
    \item Choose $k \cdot N$ states or transitions to give reward to (where $N$ is the total number of states or transitions).
    \item Sample a reward function from $U[-1, 1]$ IID, or generally from $\zeta_r$, over these states/transitions.
    \item Sample a value $\epsilon$ controlling optimality.
    \item Sample an $\epsilon$-greedy optimal policy.
\end{enumerate}

For optimal policies, UPS consists of step 5 (for a zero reward function), URS consists of steps 3 and 5, and USS consists of steps 1-3 and 5. For suboptimality, add step 4.

\subsubsection{Dense and Sparse Loss Function}
\label{Loss Funcs}

The dense loss function used in our model is based on the Cross-Entropy Loss, which is commonly employed for classification tasks. The function is defined as follows:

\begin{equation}
\mathcal{L}_{dense}(\mathbf{y}, \mathbf{\hat{y}}) = - \frac{1}{N} \sum_{i=1}^{N} \sum_{c=1}^{C} y_{i,c} \log(\hat{y}_{i,c}),
\end{equation}
\label{equation:dense}

where \(\mathbf{y}\) represents the ground truth labels, \(\mathbf{\hat{y}}\) represents the predicted logits from the model, \(N\) is the number of samples, and \(C\) is the number of classes. In this context, \(y_{i,c}\) is a binary indicator (0 or 1) if class label \(c\) is the correct classification for sample \(i\), and \(\hat{y}_{i,c}\) is the predicted probability (logit) of sample \(i\) being in class \(c\).

The sparse loss function is designed to focus on a subset of logits and labels, specifically a random subset of \(n\) tokens. This can be useful in scenarios where the model needs to pay particular attention to recent outputs. The function is defined as follows:

\begin{equation}
\mathcal{L}_{\text{sparse}}(\mathbf{y}, \mathbf{\hat{y}}, n) = \mathcal{L}_{CE}\left(\mathbf{y}_{n}, \mathbf{\hat{y}}_{n}\right),
\end{equation}
\label{equation:sparse}

where \(\mathbf{y}_{n}\) and \(\mathbf{\hat{y}}_{n}\) represent the ground truth labels and predicted logits for a random subset of \(n\) tokens, respectively. The cross-entropy loss \(\mathcal{L}_{CE}\) is applied to these extracted tokens. The implementation of the sparse loss function in PyTorch is as follows:

In this implementation, \texttt{number\_logits} specifies the number of tokens to consider. The function \texttt{extract\_last\_n\_tokens} extracts the last \(n\) tokens from both the model outputs and the labels. The tensors are then reshaped using the \texttt{view} method to ensure they are in the correct format for the cross-entropy loss function, \texttt{torch.nn.CrossEntropyLoss()}.

\subsection{Speculation}
\label{Speculation}

In this section, we discuss speculative impacts of our work on AI alignment and AI safety.

One of the most difficult parts of AI alignment is that, if you already have a misaligned optimizer trying to deceive a training process (\cite{hubinger2019risksfrom}, \cite{carlsmith2023scheming}), it is very difficult to detect or align it since it is actively scheming against your techniques. A strong optimizer of a misaligned utility function would want to protect itself from being modified or being interpretable, possibly leading to steganography (\cite{dewitt2023perfectly}) to prevent interpretable natural language outputs, or phenomena like reward hacking (\cite{ngo2022deeplearning}). By contrast, if you are able to build a highly capable AI that is not goal-directed, perhaps using a dense reward signal, then you avoid these failure modes entirely.

Even if this AI is not competitive with agentic AIs, it may make the alignment problem substantially easier. For instance, we can conceptualize the pre-training process of a transformer as providing a very dense reward signal: instead of simply providing a scalar update value, pre-training specifically updates a transformer towards certain completions, making it suitable for the bulk of compute used in the training of current LLMs. If this is true, then we can train LLMs to understand human values and ontology in the sense that it can follow directions and do what a user intends \textit{from natural language instructions alone}, without it being goal-directed. Then, if we want to agentize the LLM through, for instance, RL, we do not need a reward signal that robustly describes human preferences in all situations; we only need to provide good enough reward or prompting for it to do what we want, since it already understands what we want it to do. This substantially simplifies the alignment problem (\cite{valuemisspecification}).

More generally, as discussed in the introduction, catastrophic failure modes like goal misgeneralization and deception rely on the AI being well-modeled as being goal-directed or agentic for some definition of those terms. Proceeding with empirical work without a clear theoretical understanding of the failure mode we are trying to prevent can lead to misinterpreting results and faulty assumptions, which is suboptimal when the cost of failure could be massive in scope. Being able to apply theoretical defintions of goal-directedness to real-life models gives future experiments better grounding to attack goal misgeneralization and deceptive alignment directly, allowing for more rigorous and scientific work on prosaic alignment of current frontier models.

We hope that, with much further refinement and iteration, future methods similar to ours can determine whether current deep learning techniques (e.g. transformer pre-training, fine-tuning, RLHF, etc.) on commonly used datasets tend to form coherent optimizers that are relevant to the aforementioned catastrophic AI risk scenarios involving goal misgeneralization and deception.

\subsection{Further background}
\label{Further background}

Another relevant definition has been written by \cite{kosoydefinition} for the ``agency” of a policy in her learning-theoretic agenda. Kosoy's definition generalizes our uniform reward prior to some Solomonoff prior (\cite{SOLOMONOFF19641}), only relying on the utility of a policy with respect to some utility function and simplicity priors $\zeta$ and $\xi$ (which can be generalized to any prior over the space of policies and reals respectively). \footnote{In this section, we will discuss different definitions that refer to the \textit{coherence} or \textit{agency} of a policy or system instead of its goal-directedness. These terms are very related, and different authors use them differently; roughly speaking, coherence refers to an agent's ability to not ``contradict" itself in pursuit of some goal (\cite{vnmtheorem}), while agency is a broader term borrowing various connotations from psychology and economics (\cite{litreview}). When we refer to our model, we will only discuss the \textit{goal-directedness} of a system or policy.} Rather than using sparsity over the environment, Kosoy weighs by the Kolmogorov complexity of the utility function (\cite{Kolmogorov1998OnTO}). However, computing this definition requires taking the maximum of a function over all utility functions $U$ and universal Turing machines $M$ respectively. Meanwhile, \cite{wentworthcoherence} considers a policy coherent \textit{in the long-term }if it does not contradict itself, i.e. if there exists a value function that fits the policy and satisfies the Bellman equations with zero payoff. 

Other definitions of goal-directed behavior place more emphasis on whether a model has agentic \textit{structure}, such as a ``goal-slot" for which a model optimizes directly (\cite{hubinger2019risksfrom}). Related to this is the idea that goals should be ``concepts" in an agents' world-model (\cite{litreview}), or that agents should share human-like agentic characteristics like self-awareness or planning (\cite{agisafetyprinciples}, section 3.2). However, Ngo (Section 3.1) also discusses the problems with finding an explicit representation of a goal or a search algorithm in \cite{hubinger2019risksfrom}, and in practice more fundamental concepts such as \textit{features} remain preliminarily defined in neural networks (\cite{featuresdef}).

Our methodology was also substantially inspired by \cite{turner2023optimal}, which studies the properties of optimal policies under MDPs. They find that certain properties and symmetries of an MDP lead to power-seeking behavior by optimal policies. Specifically, for any state $s$, discount rate $\gamma$, and distribution of reward functions $D_{\text{bound}}$ with some bounding conditions, then POWER is defined as

\begin{equation}
\text{POWER}_{\mathcal{D}_{\text{bound}}} (s, \gamma) = \frac{1 - \gamma}{\gamma} \mathbb{E}_{R \sim D_{\text{bound}}} [V^{*}_{R}(s, \gamma) - R(s)]
\end{equation}

$V^{*}_R(s, \gamma)$ refers to the optimal value of a state, or the value of a state given an optimal policy over a reward function $R$. We might then say that POWER measures the expected optimal value of a state over all relevant reward functions. Then, action $a$ is more power-seeking than $a’$ when the expected POWER of $a$ is greater than the expected POWER of $a’$. We borrow intuitions from \cite{turner2023optimal} about properties of MDPs that are correlated with optimality (and by extension POWER-seeking), like 1-cycles, loops, and the ``optionality” of nodes in deterministic MDPs. Intuitively, policies sampled from URS may be more likely to ``explore” the graph of states to find a particularly high-reward group of states, thus resulting in a policy that takes longer before it starts looping between states (assuming policy invariance across time-steps). URS-sampled policies, if power-seeking, may also tend to avoid 1-loops (actions that take an agent from a state to itself). \cite{farquhar2022pathspecific} also approach the problem of agency through MDPs with specific conditions.

There has also been substantial work on \textit{causal} perspectives to agency or goal-directedness, arguing that an AI is agentic when it would act differently if its actions had different consequences (\cite{KENTON2023103963}). Kenton et al. use causal diagrams to model this differential decision-making, testing how general the AI's algorithm is and how well it can adapt to unforeseen circumstances.

Finally, some researchers working on AI alignment believe that intelligence and capabilities are inherently tied with ``coherence” (\cite{yudkowsky2017coherent}), and thus any sufficiently capable AI will approximately be a coherent utility function maximizer. Other researchers strongly disagree (\cite{nocountingargs}). There is also empirical evidence that this sort of optimization does not occur naturally in deep neural networks. For instance, \cite{veit2016residual} showed that residual networks can be modeled as having short distinct ``paths" that are relatively independently updated during training; longer paths do not receive significant gradient. If the cognition inside neural networks is modular and shallow, it seems difficult to implement long chains of if-then reasoning that is required for search. \cite{spectralbias} also shows that deep learning is biased towards low-frequency functions, which contradicts the idea of a mesa-optimizer that can arbitrarily change its behavior based on its ``goal-slot".

\subsubsection{Related fields}
\label{Related fields}

In fields like decision theory, philosophy (especially epistemology), and economics, defining how to characterize \textit{agency} or goal-directedness is a fundamental problem in the field. In decision theory and economics, there are long-standing \textit{coherence theorems} that prove that an agent's behavior can be fully (\cite{vnmtheorem}) or partially (Savage's theorem \cite{savage54}, Bolker-Jeffrey theorem \cite{Bolker-Jeffrey}) explained as preferring outcomes highly rated with some utility function, assuming some axioms about said preferences. There have also been attempts to apply philosophical preference utilitarianism in toy settings (\cite{Oesterheld_2015}). However, there are situations where expected utility maximization is unintuitive (Allais paradox, St. Petersburg paradox; see \cite{Allais1990}) or where one’s preferences can be incomplete, and expected utility can trivially describe any behavior if applied to trajectory or action histories. Incomplete and incommensurable preferences thus present a challenge for utility models of human preferences. Expected utility maximization also usually assumes a dualist, causal framework between agent and environment where the agent's decisions are ``uncaused", but in practice the agent is part of and can be influenced by the environment. Building non-dualist models of agency (e.g. Bayesian decision theory, see \cite{EDTintro}) is a relevant direction of future work.

\subsection{Better baselines}
\label{Better baselines}

One problem we might face in following the above strategy: what if it is computationally too easy to distinguish policies sampled via UPS from policies sampled via USS? If so, binary classifier values of $G(\pi_0)$ might cluster near 0 or near 1, leading to numerical problems. In other words, for highly coherent policies, UPS is a very poor baseline to compare USS against. So what if we used a series of baselines for training classifiers instead? For example, we could calculate goal-directedness as:

\[
G(\pi_0) = \frac{P(\pi=\pi_0 | USS)}{P(\pi=\pi_0 | UPS)} = \frac{P(\pi=\pi_0 | USS)}{P(\pi=\pi_0 | URS)} \frac{P(\pi=\pi_0 | URS)}{P(\pi=\pi_0 | UPS)}
\]

This would be useful given the assumption that URS is a good baseline for USS, and UPS is a good baseline for URS. We might also be interested in other sampling strategies which are, intuitively speaking, ``somewhere between” USS and UPS. One possibility is \textit{uniform value sampling} (UVS). By UVS I mean the following procedure:

\begin{enumerate}
    \item Sample a random value function by assigning every state a value from U(-1,1).
    \item Sample a random reward function which is consistent with that value function. (Note that a) there is some state-action reward function consistent with any value function; and b) for any given value function, most state-action reward functions are \textit{not} consistent with it.)
    \item Sample an optimal policy for that reward function.
\end{enumerate}

One of the benefits of using UVS as an intermediate baseline is that knowing the value function makes it very easy to translate a reward function to an optimal policy. Another possible intermediate baseline is \textit{uniform trajectory sampling}—sampling a given trajectory (or set of trajectories), then sampling a reward function consistent with that trajectory being optimal, then sampling an optimal policy for that reward function.

\subsection{Mathematical equivalences}
\label{Mathematical equivalences}

\subsubsection{Comparison to Orseau}

We compare our method to \cite{agentsanddevices} via the following transformation. Suppose instead of uniform, IID reward functions (in our terminology) or utility functions (in their terminology), we weigh all utility functions $u$ by some $w_u$. We can define $w_u := \frac{1}{|\mathcal{U}|}$ where $|\mathcal{U}|$ is the number of utility functions if finite, or we can attach a speed prior by defining $w_u = 2^{-\lambda l(u)}$ for some constant $\lambda$ where $l(u)$ is the \textit{description length} of the utility function according to some Turing-complete reference machine. 

They then generalize from deterministic to probabilistic policies by defining $\pi_{u, \epsilon}(y_{<T} | x_{<T})$ to be the probability of an $\epsilon$-greedy (i.e. almost optimal) policy taking the trajectory of actions $y_{<T}$ conditional on the observations $x_{<T}$ and the utility function $u$. They then obtain a mixture of probabilities via a weighted sum over all goals $g$:
\[
M_g(y_{<T} | x_{<T}) := \sum_{u \in \mathcal{U}} w_u \pi_{u, \epsilon}(y_{<T} | x_{<T})
\]
Meanwhile, we calculate $M_d(y_{<T} | x_{<T}) := \sum_{d \in \mathcal{M}_d} w_d d(y_{<T} | x_{<T}$ by taking some prior distribution $w_d$ over all possible ``systems" $d$ that, similar to $\pi$, return a probability distribution of actions over observations. We roughly approximate this using uniform policy sampling, but their definition seems more elegant. Finally, like our method, we calculate the probability that a system (or policy) is an agent or a device using Bayes' rule:
\[
P(\text{agt} \mid y_{<t}, x_{<t}) = \frac{M_g(y_{<t} \mid x_{<t})}{M_d(y_{<t} \mid x_{<t}) + M_g(y_{<t} \mid x_{<t})}
\]
\[
P(\text{dev} \mid y_{<t}, x_{<t}) = \frac{M_d(y_{<t} \mid x_{<t})}{M_d(y_{<t} \mid x_{<t}) + M_g(y_{<t} \mid x_{<t})}
\]

\subsubsection{Comparison to Kosoy}

The method in \cite{kosoydefinition} is somewhat similar to Orseau, except outputting a scalar value instead of a probability. Specifically, we equate $l(u) = K(u, M)$ as the Kolmogorov complexity/bits required to specify a utility function $u$ given some Turing-complete machine $M$. Then over all utility functions $u \in U$ and Turing machines $M$, Kosoy defines
\[
\text{g}(\pi) := \sup_{u, M} \left( \min_{\pi' : \mathbb{E}_{\xi_M \pi'}[u] \geq \mathbb{E}_{\xi_M \pi}[u]} K(\pi') - K(u, M) \right)
\]
In other words, Kosoy finds the pair $(u, M)$ such that the minimum complexity difference between the policy (or any policy that is better than it at achieving high utility given $u$) and the utility function is maximized. Roughly speaking, this is the complexity of the ``simplest" $u$ that ``describes" $\pi$ well. Maximimally agentic policies like AIXI (\cite{hutter2000theory}) receive $g(\pi) = +\infty$, while fundamentally ``complex" or ``random" policies (like our UPS-sampled policies) receive near-zero $g$.

\subsubsection{Retargetable policies}

Turner later extended his work to policies with retargetable cognition (\cite{turner2022parametrically}). As another intuition pump, if a policy $\pi$ is optimal for many reward functions, then it tends to be retargetable over many permutations of a reward function. Hence $P(\pi = \pi_0 | URS)$ measures the distribution of retargetability, which seems useful.

\subsection{Miscellaneous theoretical arguments}
\label{Miscellaneous theoretical arguments}

\subsubsection{Against myopia}

One particular objection that some may have about our definition is that, even if coherent policies meaningfully tend to maximize reward functions, those reward functions may in practice be ``low-impact”, and thus not matter for AI risk. One example is the concept of a ``myopic” AI, which is only goal-directed within a small time-frame, and hence cannot affect the world in ways we would consider dangerous. We give preliminary empirical evidence that coherent policies tend to pursue long-term reward (at least with a high enough discount rate, e.g. 0.9). We can also provide a loose argument that myopic policies will tend to have low goal-directedness.

Suppose you have a policy $\pi$ that is myopic at a state $s$. Then we can model the policy as taking the action $a$ with the highest expected next-step reward $\mathbb{E}_{s’ \in S} [R(s, a, s’)]$, which given that the MDP is deterministic, equals some $R(s, a)$. If this policy is optimal for this reward function, then $R(s, a)$ will be very high, and there will be many policies that are also myopic in taking action $a$ at state $s$, and are also optimal for $R$ at $s$. But then $P(\pi = \pi_0 | URS)$ will be low, as $\pi$ is only one of many policies taking the same action at $s$. Therefore, its goal-directedness will also be low; this argument works similarly for $P(\pi = \pi_0 | USS)$.

\subsection{MDP experiment terminology}
\label{MDP experiment terminology}

\begin{enumerate}
    \item \textbf{(P).} One ``brute force” method is by joining the (tuple) optimal policy $\pi_0$, flattened transition function, and discount rate into a 1-dimensional vector. This in theory contains all the information about the MDP and $\pi_0$ that we can provide, but in practice needs more processing before it can be classified. (Again, a more principled approach would likely involve some kind of graph neural network.)
    \item \textbf{(LL).} Given that $\pi$ is deterministic in a deterministic MDP, let $\pi(s_t) = s_{t+1}$ for all $t \geq 0$. Then eventually $s_{t_1} = s_{t_2}$ for some $t_1$ and $t_2$, at which point the policy will be in a ``loop". Thus another possible set of features is, for every state $s_0$, measuring \textit{how long} it takes for the optimal policy $\pi_0$ to reach a loop when starting from $s_0$, and how long the loop itself is. We can think of optimal policies as implementing an \textit{explore and exploit} dynamic: navigating to a particularly high-reward area of the MDP, and then looping through that area to maximize reward indefinitely. Intuitively, a policy that takes longer to reach a stable loop can access more of the MDP and can thus reach higher-reward areas, while a policy that takes a bigger loop can incorporate more reward into the loop.
    \item \textbf{(O+S).} Finally, if optimal policies are “power-seeking”, then we can try using correlates of POWER (\cite{turner2023optimal}). Specifically, we use the sum of the number of \textit{out-arrows}, or directly reachable states, from a given state $s$ for all $s$ that an optimal policy $\pi_0(s_0)$ reaches, and a binary value representing whether $\pi_0(s_0) = s_0$ self-loops indefinitely.
\end{enumerate}

\subsection{Miscellaneous MDP results}
\label{Miscellaneous MDP results}

\begin{figure}
  \centering
  \begin{subfigure}[b]{0.5\textwidth}
    \centering
    \includegraphics[width=\linewidth]{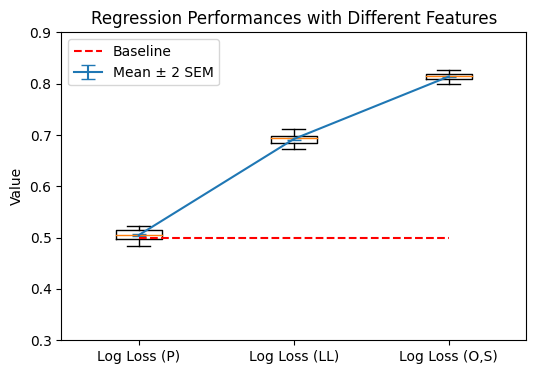}
    \caption{Accuracy graph}
    \label{fig:sub3}
  \end{subfigure}%
  \begin{subfigure}[b]{0.5\textwidth}
    \centering
    \includegraphics[width=\linewidth]{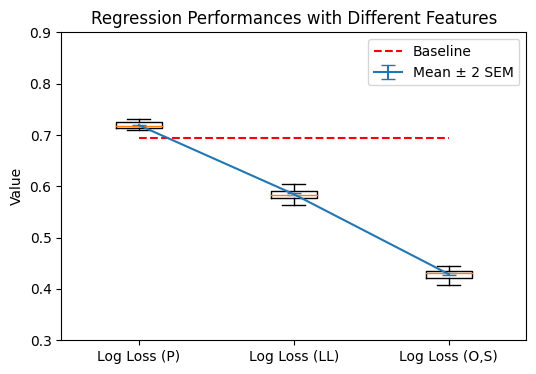}
    \caption{Loss graph}
    \label{fig:sub4}
  \end{subfigure}
  \caption{Accuracy and loss when passing in three different sets of features into the logistic classifier for predicting $\frac{P(\pi = \pi_0 | USS)}{P(\pi = \pi_0 | UPS)}$.}
  \label{fig:main2}
\end{figure}

\begin{figure}
  \centering
  \begin{subfigure}[b]{0.5\textwidth}
    \centering
    \includegraphics[width=\linewidth]{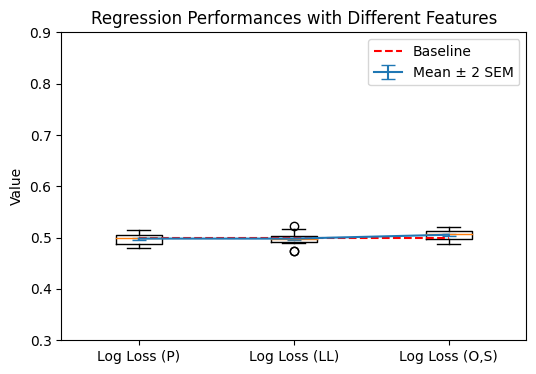}
    \caption{Accuracy graph}
    \label{fig:sub1appx}
  \end{subfigure}%
  \begin{subfigure}[b]{0.5\textwidth}
    \centering
    \includegraphics[width=\linewidth]{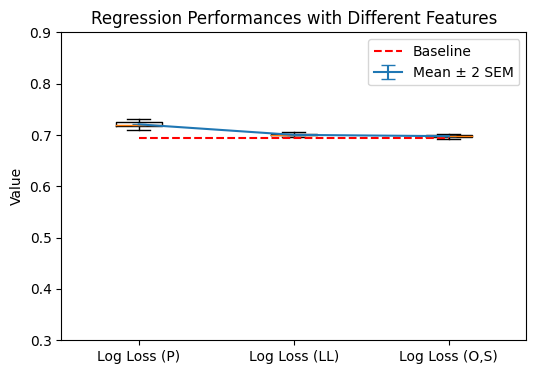}
    \caption{Loss graph}
    \label{fig:sub2appx}
  \end{subfigure}
  \caption{Accuracy and loss when passing in three different sets of features into the logistic classifier for predicting $\frac{P(\pi = \pi_0 | USS)}{P(\pi = \pi_0 | URS)}$.}
  \label{fig:main1appx}
\end{figure}

We performed additional tests on goal-directedness. When we try to build a classifier for the $\frac{P(USS)}{P(URS)}$ definition of goal-directedness, we find that our current classifier architectures and features are insufficient, as shown in Figure \ref{fig:main1appx}. Some other results:

\begin{enumerate}
    \item Less structured MDPs, such as MDPs where the transition probability distribution for each $T(s, a)$ (for any state $s$ and action $a$) were IID randomized via Dirichlet distribution, tended to be harder to build a classifier for. Indeed, when we sampled from this set of MDPs, randomized the reward function $10^4$ times, and then calculated the optimal policy via value or policy iteration for each reward function, we found that the resulting distribution of optimal policies was roughly uniform (the mode policy occurred 1-3 times), and did not become less uniform with increased sparsity. This would make it harder to distinguish optimal policies from uniformly randomly generated policies. We found a similar, if slightly weaker, result for random deterministic MDPs (where $T(s, a)$ is 1 for some random $s’$ and 0 for all other states).
    \item Looking at the logistic coefficients of the logistic when using self-loops and out-arrows individually as features, we found that more out-arrows correlated with a greater chance of a policy being sampled from URS/USS rather than UPS, while more self-loops correlated with a lesser chance. This matches, with weak confidence, what we would expect if ``coherent” policies optimal for some reward function tended to preserve optionality, which was hypothesized in \cite{turner2023optimal}.
\end{enumerate}

\begin{figure}
  \centering
  \begin{subfigure}[b]{\textwidth}
    \centering
    \includegraphics[width=\linewidth]{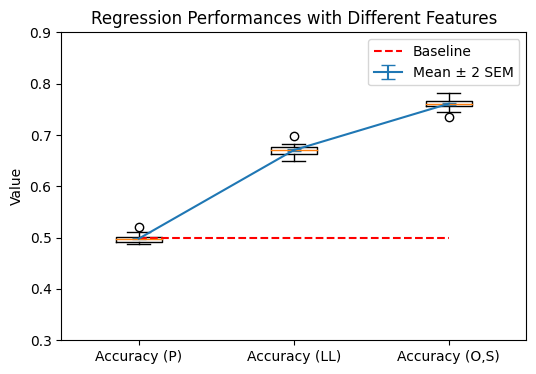}
    \caption{Accuracy graph}
    \label{fig:sub1}
  \end{subfigure} \\
  \begin{subfigure}[b]{\textwidth}
    \centering
    \includegraphics[width=\linewidth]{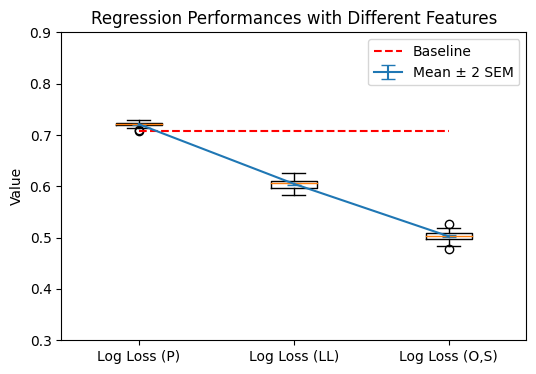}
    \caption{Loss graph}
    \label{fig:sub2}
  \end{subfigure}
  \caption{Bigger version of figure \ref{fig:main1}}
  \label{fig:blownup1}
\end{figure}

\begin{figure}
  \centering
  \begin{subfigure}[b]{\textwidth}
    \centering
    \includegraphics[width=\linewidth]{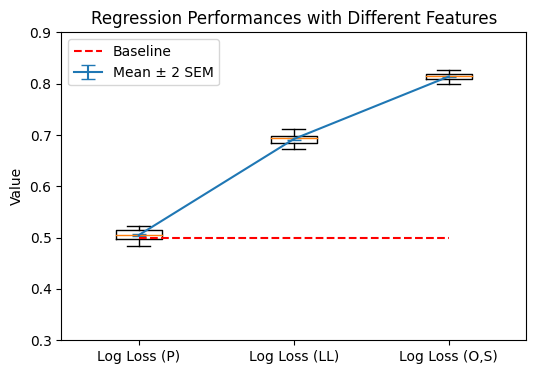}
    \caption{Accuracy graph}
    \label{fig:sub3}
  \end{subfigure} \\
  \begin{subfigure}[b]{\textwidth}
    \centering
    \includegraphics[width=\linewidth]{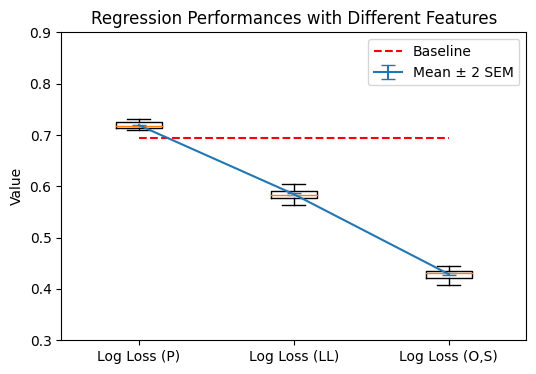}
    \caption{Loss graph}
    \label{fig:sub4}
  \end{subfigure}
  \caption{Bigger version of figure \ref{fig:main2}}
  \label{fig:blownup2}
\end{figure}

\subsection{RL results}
\label{Miscellaneous RL results}

We now present experiments in a more complex environment where it is intractable for optimal policies to be found via value or policy iteration. We approximate sampling near-optimal policies via USS or URS by training policies on their respective reward functions. We then train a classifier using these policies \footnote{We found that the elements of Q-tables generated via our RL implementation of USS tended to be much lower in magnitude than in Q-tables generated via URS. It seems unclear whether this actually correlates with increased goal-directedness or coherence, so our classifier could thus overfit when passing in Q-tables. We thus stuck to passing in only the policy into the FCNN and GNN; any patterns the classifier found between USS and URS policies should point to differences resulting from training policies in general with a sparse versus a dense reward function, i.e. what we want to measure.} and run the classifier on test cases to test if it is measuring goal-directedness properly. The purpose of these experiments is to see whether our definition of goal-directedness can be measured cheaply and effectively in complex environments.

In the Taxi-v3 environment in OpenAI's Gym (\cite{brockman2016openai}), we trained 50 Q-tables with reward functions sampled from USS for 10,000 episodes each using Q-learning (\cite{qlearning}), and 50 trained similarly with reward functions from URS. We used state-based rewards for URS and USS, and a sparsity value of $\epsilon = 0.99$. We generated a policy greedy in value for each Q-table, then labeled each policy with which method it was generated by, and trained an ensemble of 40 binary classifiers using a graph convolutional neural network (\cite{zhou2021graph}) (GCN) defined over the state and action space of the environment.

We found that a GCN classifier could effectively predict the labels (< 0.2 test loss with enough epochs). We also used $\frac{P(USS)}{P(URS)}$ as a benchmark instead of $\frac{P(USS)}{P(UPS)}$, intuitively measuring the sparsity of a policy's reward function, for an intermediate baseline, as described in section \ref{Better baselines}. We then wanted to test whether these classifiers actually identified what we intuitively think of as "coherent" policies. However, we found inconclusive results in this regard; see section \ref{Generalization tests}.

\begin{figure}
  \centering
  \begin{subfigure}[b]{0.5\textwidth}
    \centering
    \includegraphics[width=\linewidth]{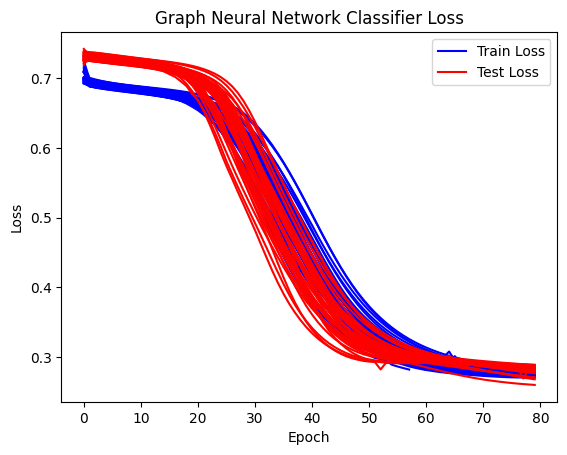}
    \caption{For $\frac{P(URS)}{P(UPS)}$}
    \label{fig:lossapx1}
  \end{subfigure}%
  \begin{subfigure}[b]{0.5\textwidth}
    \centering
    \includegraphics[width=\linewidth]{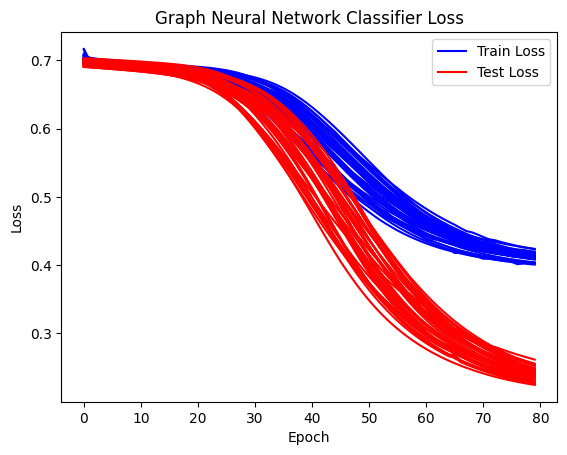}
    \caption{For $\frac{P(USS)}{P(URS)}$}
    \label{fig:lossapx2}
  \end{subfigure}
  \caption{Train and test loss of an ensemble of 40 GCN classifiers given a binary classification task with datasets of size 50. Learning rate = 0.003, test/train ratio = 0.2}
  \label{fig:main1appx}
\end{figure}

\subsubsection{Generalization tests}
\label{Generalization tests}

We wanted to test how well our GNN classifiers generalize, meaning how well our GNN classifiers are able to predict goal-directedness for policies outside of the training dataset. We thus generated an ensemble of successful GNN classifiers, and passed in different sets of policies as test cases:

\begin{enumerate}
    \item \textbf{Taxi Q-table Policies.} We generate Q-tables using Q-learning on Taxi with the normal reward function shown \href{https://gymnasium.farama.org/environments/toy_text/taxi/}{here}, then generate greedy policies for each Q-table. Intuitively, policies subject to higher optimization pressure for some reward function should be rated high on the classifier for $\frac{P(URS)}{P(UPS)}$, where a rating of 1 means URS and a rating of zero means UPS. The normal reward function for Taxi is somewhat sparse (+20 if delivering passenger to correct location, -10 if executing "pickup" and "drop-off" actions illegally, -1 otherwise), so that may also lead to a higher coherence score.
    \item \textbf{MCTS policies.} Policies generated from conducting MCTS search (\cite{mcts_wiechowski_2022}) for 1000 iterations, using the Q-tables from our first set of policies (our previous RL training on Taxi) as rollout policies. The MCTS algorithm is often cited (\cite{hubinger2019risksfrom}) as an example of what a goal-directed model will internally emulate, so our hypothesis is that MCTS-generated policies will be more coherent and goal-directed according to our classifiers.
    \item \textbf{Non-stable-state policies.} A la \cite{turner2023optimal}, given state-based rewards, near-optimal policies should prefer to stay in high-reward states. Thus, for each Taxi policy we trained in our first dataset, we randomly (with a $50\%$ IID chance) switch the action at each state $s$ where the policy stays at $s$ to actions that move to a different state.
\end{enumerate}

\begin{figure}
  \centering
  \begin{subfigure}[b]{0.48\textwidth}
    \centering
    \includegraphics[width=\linewidth]{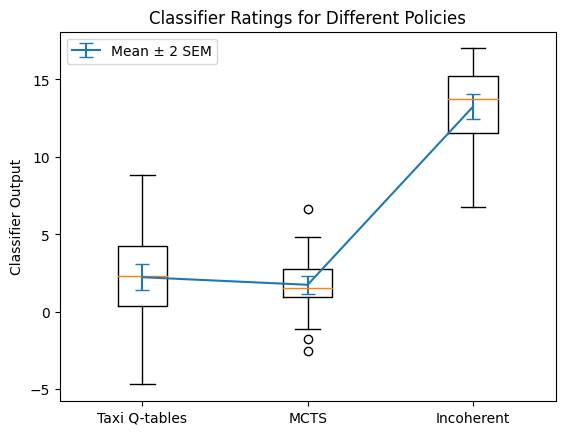}
    \caption{Logit classifier ratings for the $\frac{P(URS)}{P(UPS)}$ classifier. Note that, in order to avoid numerical issues with classifying incoherent policies, we train the Taxi Q-tables for only ten episodes each.}
    \label{fig:sub5}
  \end{subfigure} \hfill
  \begin{subfigure}[b]{0.48\textwidth}
    \centering
    \includegraphics[width=\linewidth]{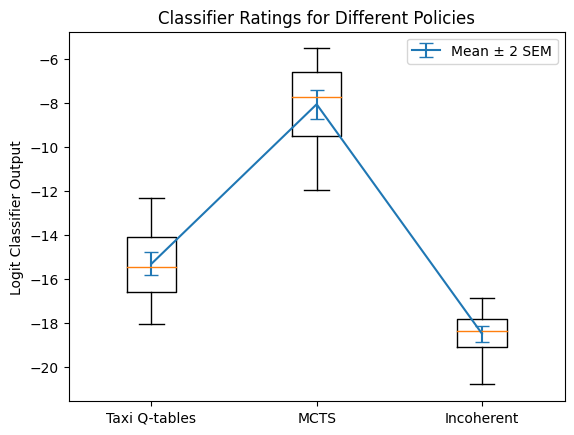}
    \caption{Logit classifier ratings for the $\frac{P(USS)}{P(URS)}$ classifier. Here, we train the Taxi Q-tables for 20,000 episodes each.}
    \label{fig:sub6}
  \end{subfigure}
  \caption{Logit binary classifier ratings given the datasets described previously. All error bars shown in the RL experiments assume a normal distribution and show the two-sigma error of the independent sampling of each value graded by 30-40 independently generated classifiers.}
  \label{fig:main3}
\end{figure}

Our results are shown in Figure \ref{fig:main3}. If our classifier rates $p=1$, then it believes that $P(USS) = 1$ and $P(URS) = 0$ if classifying $\frac{P(USS)}{P(URS)}$ or $P(URS) = 1$ and $P(UPS) = 0$ if classifying $\frac{P(URS)}{P(UPS)}$. If our classifier rates $p=0$, then it believes the opposite in both cases. We feed our classifier ratings through the logit function $l(p) = \ln{\frac{p}{1-p}}$ for better analysis and visualization.

The Taxi Q-tables were classified as strongly URS, and remained that way for a non-trivial number of trained episodes (>10). Although the magnitude varied by test, the MCTS policies were graded to be roughly as coherent as the policies from the Taxi Q-tables, but with the logit classifier rating having a lower magnitude. We speculate that this is the result of some misgeneralization from the training data because of the MCTS process, resulting in the classifier not having predictions as confident as for policies generated from Taxi Q-tables. Finally, policies with fewer 1-cycle states were graded as strongly URS, more so than the Taxi Q-tables, which was surprising.

\begin{figure}
  \centering
  \begin{subfigure}[b]{0.48\textwidth}
    \centering
    \includegraphics[width=\linewidth]{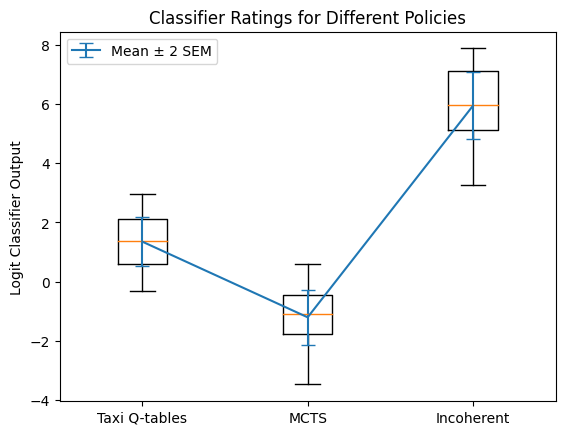}
    \caption{Logit classifier ratings for the $\frac{P(URS)}{P(UPS)}$ classifier. Note that, in order to avoid numerical issues with classifying incoherent policies, we train the Taxi Q-tables for only ten episodes each.}
    \label{fig:sub5}
  \end{subfigure} \hfill
  \begin{subfigure}[b]{0.48\textwidth}
    \centering
    \includegraphics[width=\linewidth]{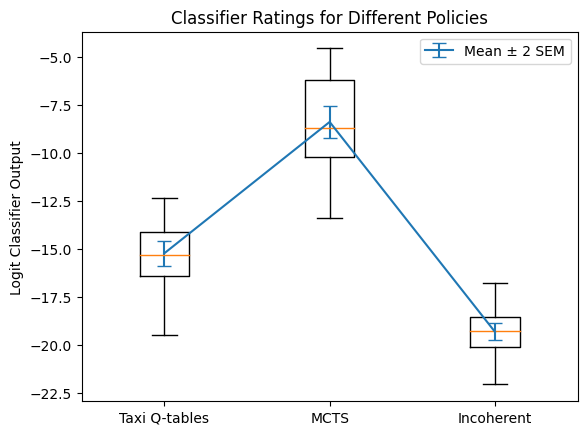}
    \caption{Logit classifier ratings for the $\frac{P(USS)}{P(URS)}$ classifier. Here, we train the Taxi Q-tables for 20,000 episodes each.}
    \label{fig:sub6}
  \end{subfigure}
  \caption{Logit binary classifier ratings with lower binary classifier losses (0.5 on average)}
  \label{fig:main5appx}
\end{figure}

We note that considerable effort was put to make sure that no biases were unintentionally left in the datasets, potentially leading to the classifiers measuring a confounding factor. For example, if the classifiers overfit to inductive biases in the training process, then it would be measuring ``how much" the policies were trained instead of goal-directedness itself, which is undesirable. To combat this, we incorporated policies trained with MCTS for 1000 iterations into half of the USS and URS policy datasets. However, we did see some negative evidence in previous experiments; these are also listed in the appendix. These results are also sensitive to hyperparameters; $\frac{P(URS)}{P(UPS)}$ classifiers with greater test loss, for instance, tended to rate MCTS with lower goal-directedness than the Taxi Q-table policies they were derived from (Figure \ref{fig:main5appx}). We can surmise that that the high sparse reward in Taxi (when the passenger is successfully dropped off) can only be achieved once, while given randomly sampled sparse reward functions over states, policies can loop around and repeatedly earn high rewards until the maximum number of states is reached. Thus Taxi policies functionally act as if they are in URS; we note that this implies that certain features in the environment are required or helpful for policies to exhibit goal-directed behavior, as \cite{turner2023optimal} also shows for power-seeking behavior. For a similar reason, policies that are changed to not have cycles (while also still pursuing high-reward states), like Taxi policies, are similarly graded URS. However, we note that the above explanation, along with the RL generalization experiments in general, are incomplete/inconclusive, which is why we acknowledged as such in this paper and put those experiments in the appendix.

More generally, coming up with good, tractable ground truth examples to check that our classifiers measure an already-nebulous concept like goal-directedness is difficult. Thus, we should not put too much weight on positive or negative results on these specific benchmarks. We thus see some evidence that the predictions of our classifiers match our intuitions around goal-directedness, which lends evidence to the claim that our classifiers are identifying goal-directedness properly.

\subsubsection{Hyperparameters}

Below we list additional data from our RL experiments. In figure \ref{fig:main1appx}, we graph the loss of a sample batch of 40 graph convolutional network (GCN) classifier runs on both binary classification tasks. Each data point consisted of a deterministic greedy policy generated from a Q-table of dimension $(|S|, |A|) = (500, 6)$. Each table was trained with learning rate $\lambda = 0.01$ and discount rate $\gamma = 0.99$ for some number of episodes, under standard Q-learning (\cite{qlearning}) with exploration factor $\epsilon$ decaying from 1.0 to 0.01 over the first half of the total number of episodes. For URS and USS policies, each Q-learning run also had a state-based reward function, where $R(s) \sim U[-1, 1]$ I.I.D. in the case of URS or for a $1-k=0.01$ proportion of the states of USS policies. For the other $k \cdot 100\% = 99\%$ states, $R(s) \simeq 0$ (technically $U[10^{-9}, 10^{9}]$ to avoid issues with many possible similar-performing policies).

After generating these deterministic policies, we trained a 2-layer, 16-node-wide GCN with a singular output, parsed through a sigmoid function to create a binary prediction. This training process had a learning rate between 0.01 and 0.1 depending on the run, but most commonly (and in Figure \ref{fig:main1appx}) had $\lambda = 0.03$, along with a rate decay of $5 \cdot 10^{-4}$ over 80-150 epochs. Classifiers that ended up with a test loss of less than 0.6 (i.e. better than average) were added to the ensemble for testing.

\begin{figure}
  \centering
  \begin{subfigure}[b]{0.5\textwidth}
    \centering
    \includegraphics[width=\linewidth]{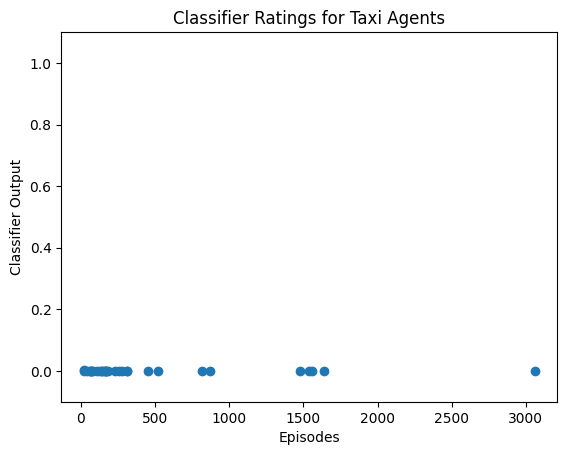}
    \caption{For 0-3000 episodes}
    \label{fig:lossapx1}
  \end{subfigure}%
  \begin{subfigure}[b]{0.5\textwidth}
    \centering
    \includegraphics[width=\linewidth]{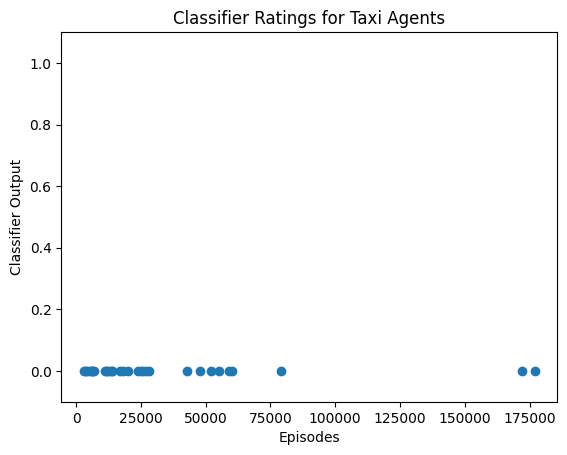}
    \caption{For 0-175000 episodes}
    \label{fig:lossapx2}
  \end{subfigure}
  \caption{$\frac{P(USS)}{P(URS)}$ classifier ratings (pulled randomly from the ensemble) of Q-tables trained on Taxi with the normal reward function for a given number of episodes. Distribution of episodes generated via \texttt{int(np.random.lognormal(3, 1)) * i} where \texttt{i} is 10 in the left graph and 1,000 in the right graph. Since all the values are near-zero, the standard deviation and error is thus also near-zero (although the distribution is not well-modeled as normal anyways).}
  \label{fig:main2appx}
\end{figure}

\begin{figure}
  \centering
  \begin{subfigure}[b]{0.5\textwidth}
    \centering
    \includegraphics[width=\linewidth]{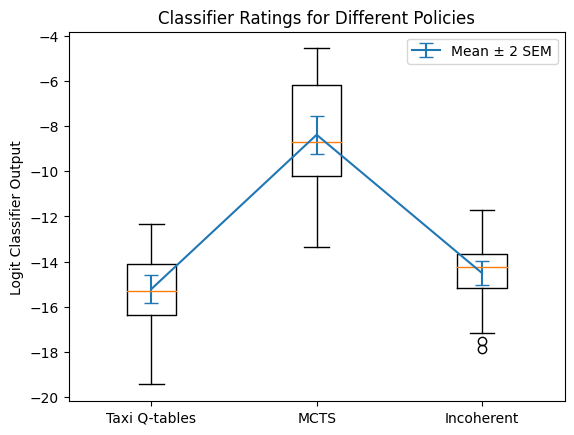}
    \caption{For $\frac{P(USS)}{P(URS)}$}
    \label{fig:lossapx1}
  \end{subfigure}%
  \begin{subfigure}[b]{0.5\textwidth}
    \centering
    \includegraphics[width=\linewidth]{images/output_USS_URS_unfixed_1.png}
    \caption{For $\frac{P(URS)}{P(UPS)}$}
    \label{fig:lossapx2}
  \end{subfigure}
  \caption{Classifier ratings with the old (in)coherence definition according to \cite{wentworthcoherence}.}
  \label{fig:main3appx}
\end{figure}

\subsubsection{Additional tests and limitations}

We ran multiple additional tests on our ensemble classifiers to see how well they generalize. We ran Taxi Q-table agents (with the default reward function) for a variable number of episodes, to see whether the $\frac{P(USS)}{P(URS)}$ ratings remained consistent. As shown in Figure \ref{fig:main2appx}, the classifier ratings remained near-zero no matter how many episodes (greater than a trivial amount, like ten) we trained the Taxi Q-table agents for. This is important because it indicates that we can train a goal-directedness classifier on policies trained with a small number of episodes, thus being computationally cheaper to generate, and have it generalize to policies trained with a larger number of episodes.

However, we had an additional test for our goal-directedness classifiers that ended up failing. According to \cite{wentworthcoherence}, we can consider the \textit{long-term} coherence of a determinstic policy in an MDP as whether there exists a value function \textit{with zero immediate payoff} that satisfies the Bellman equation (\cite{bellmaneq}) and is consistent with the policy. As such, to create an ``incoherent" dataset, we take our policies from our Taxi Q-tables and randomly switch about 70-80 actions such that the policy gets stuck in loops and is not optimal for any value function. This process should result in a set of policies with lower goal-directedness, but as shown in Figure \ref{fig:main3appx}, this process ends up ultimately giving roughly equal goal-directedness as rated by our classifiers. Indeed, note that the actions are labeled from 0 to 5, with actions 4 and 5 resulting in no change to the environment (unless some specific conditions are held, i.e. if the taxi is at the passenger location or at the destination location with the passenger onboard). When we switched \textit{all} of the actions in each policy in this third dataset such that we added one to even actions and subtracted one to odd actions (meaning that policies that stayed at a state continued to stay at a state), our goal-directedness rating was still unchanged. This gave strong evidence that our classifiers were not measuring something akin to the definition proposed by \cite{wentworthcoherence}, but instead something else (like what we measured in section 4.2, i.e. how often the policy tended to stay at a state $s$ for every possible $s$). This gives negative evidence that our classifiers are generalizing properly.

\begin{figure}
    \centering
    \includegraphics[width = 0.6\linewidth]{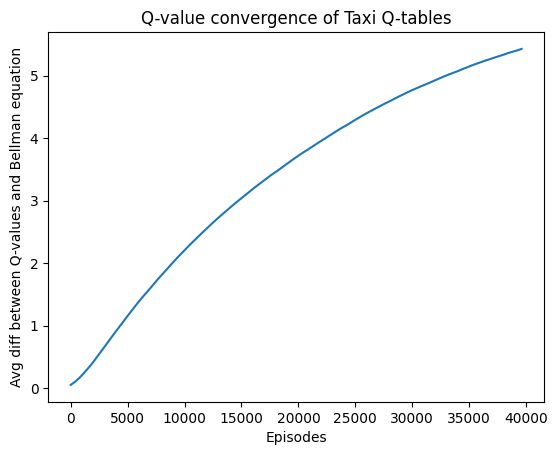}
    \caption{The average differences between each element of a Q-table and its ``ideal" value, as proposed by \cite{wentworthcoherence}, of a Taxi Q-table in training over some episodes.}
    \label{fig:main4appx}
\end{figure}

On the other hand, perhaps the definition provided by \cite{wentworthcoherence} is not a good definition for our purposes in the first place. To test this, we train a Taxi Q-table using Q-learning RL and, at intervals of 400 episodes, calculate the difference between a Q-value $Q(s, a)$ and the maximum of the Q-values at the next state $s'$ given state $s$ and action $a$:
\[
|Q(s, a) - \gamma \max_{a' \in \mathcal{A}} Q(s', a')|
\]
Theoretically, if $\gamma \approx 1$ and the relative weight of near-term rewards goes to zero, and $Q(s, a)$ is near-optimal for some utility function, then this difference should be as close to zero as possible. However, as shown in Figure \ref{fig:main4appx}, this difference actually increases with the number of episodes. In fairness, Figure \ref{fig:main2appx} shows that Taxi Q-tables are not maximally goal-directed as the number of episodes goes to infinity, instead gravitating towards being classified as URS policies, so this comparison is probably not fair. Nevertheless, this shows that modelling goal-directed policies as approximately satisfying a Bellman equation \textit{with zero payoff} is a very strong assumption that may not match what we want to describe with goal-directedness.

We also attempted to use a logistic classifier directly over the policy as a goal-directedness classifier. This did not work and ended up overfitting.

\begin{figure}
    \centering
    \includegraphics[width = 0.6\linewidth]{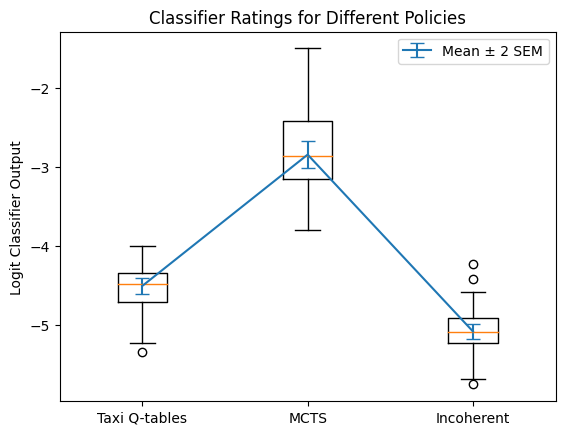}
    \caption{Logit binary classifier ratings of $\frac{P(USS)}{P(URS)}$ with $k = 0.1 \cdot N$. We see that all the policies are graded significantly closer to USS than in Figure \ref{fig:main3}.}
    \label{fig:main6appx}
\end{figure}

\begin{figure}
    \centering
    \includegraphics[width = 0.6\linewidth]{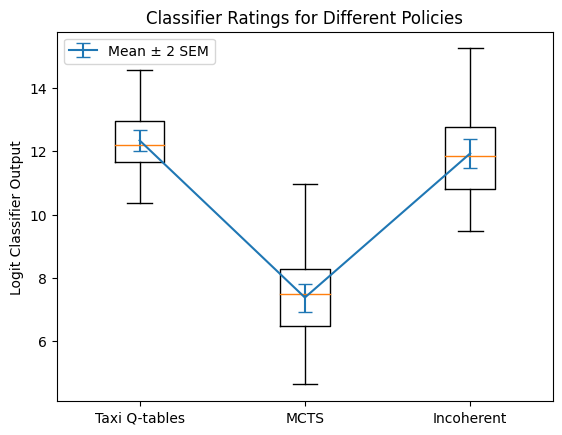}
    \caption{Logit ensemble binary classifier ratings for $\frac{P(URS)}{P(UPS)}$ using Taxi policies trained for 20,000 episodes (and MCTS and reduced 1-cycle policies using those policies).}
    \label{fig:main7appx}
\end{figure}

\subsection{Miscellaneous LLM results}
\label{Miscellaneous LLM results}

The implementation of the dense loss function in PyTorch is as follows:

\begin{verbatim}
def dense_loss(model_output, labels, n=None):
    loss_fct = torch.nn.CrossEntropyLoss()
    loss = loss_fct(model_output.logits.view(-1, model_output.logits.size(-1)), labels.view(-1))
    return loss
\end{verbatim}

In this implementation:
\begin{itemize}
    \item \texttt{model\_output.logits} are the predicted logits from the model.
    \item \texttt{labels} are the true class labels.
    \item \texttt{torch.nn.CrossEntropyLoss()} is the loss function provided by PyTorch for calculating cross-entropy loss.
    \item The \texttt{view(-1, model\_output.logits.size(-1))} operation reshapes the logits and labels to ensure they are in the correct format for the loss function.
\end{itemize}

\begin{verbatim}
def sparse_loss(model_output, labels, number_logits=10):
    n = number_logits
    # Extract the last n logits and labels
    logits = extract_random_n_tokens(model_output.logits, n)
    labels = extract_random_n_tokens(labels, n)
    
    # Flatten the tensors for cross-entropy loss calculation
    logits = logits.view(-n, logits.size(-1))
    labels = labels.view(-n)
    
    loss_fct = torch.nn.CrossEntropyLoss()
    loss = loss_fct(logits, labels)
    return loss
\end{verbatim}

We also found preliminary evidence that our sparse-1-token LLM classifier correctly predicts, with approximately $66\%$ accuracy, a model trained with direct preference optimization (\cite{rafailov2024dpo}) as sparse, confirming generalization to a frontier training technique with sparse loss signal (as DPO only backpropagates once per sequence of text) used in current research. We note that this is a preliminary tentative result. 

\subsection{Limitations and future work}
\label{subsec:limitations}

One challenge with this project is that, because the ``coherence" or ``goal-directedness" of policies are difficult to definitively categorize outside of toy examples, it is difficult to determine a ``ground truth" as to whether our classifiers are actually measuring goal-directedness or not. For example, we want our goal-directedness classifiers to not just be measuring the test performance of a policy, as this assumes that policies trained in a certain manner will be goal-directed, which is the hypothesis that we are trying to test in the first place. As our experiments get more and more complex, our metric becomes progressively more and more approximate. Stronger, more efficient classifiers will likely use interpretability techniques (e.g. \cite{nanda2023emergent}, \cite{zou2023representation}, \cite{turner2023activation}).

There is also room for progress on our theoretical model of coherence. For instance, we also define our reward functions over states in the environment, while in real life, it seems more realistic to define reward functions over human abstractions or concepts (\cite{udellshard}), i.e. patterns in the environment. Extensions to partially-observable settings utilizing belief-states instead of ``ground-truth" states would also be useful, e.g. \cite{biehl2022interpretingsystemssolvingpomdps}. Our usage of sparsity could also be replaced with some notion of compressibility, perhaps using the inductive biases of the network being studied (\cite{inductivebiases}). Finally, the model could be generalized to continuous state spaces; our current method of dealing with continuity is to discretize the space and then apply our uniform reward sampling methods, but this can fail computationally.

Finally, within the experiments that included the LLMs, there are some limitations that must be addressed. Our results show that the classifier for the dense loss function can generalize between two datasets; however, we have not shown this can generalize between different types of models. For instance, our goal-directedness classifiers may be overfitting to inductive biases in our fine-tuning optimization process in general, rather than from our fine-tuning process to any specific dataset. Expanding our train and test dataset to include neural networks trained with various hyperparameters on more datasets could strengthen our results.

\subsection{Recommendations}
\label{subsec:recommendations}

As discussed previously, goal-directedness is a crucial requirement for the most dangerous scenarios involving smarter-than-human AIs in the future. We thus recommend that more research be done into measuring goal-directedness for frontier models in realistic environments. Concretely, this could result in goal-directedness being added into evaluations of potentially dangerous new models, such as those outlined for models ASL-3 and above in Anthropic's Responsible Scaling Policy (\cite{anthropicrsp}). However, we want to emphasize that our work is quite preliminary and \textit{substantial} research needs to be done before we can form reliable goal-directedness evaluations.

More broadly, understanding goal-directedness is a bottleneck to devising clearer threat models and better solutions to preventing misaligned AIs (\cite{litreview}, \cite{agisafetyprinciples}, \cite{carlsmith2023scheming}). Thus, better measurements of goal-directedness of deep learning systems could allow future experiment to know whether they are working on goal-directed models or not, reducing assumptions. This along with more speculative impacts are discussed further in section \ref{Speculation}.

\subsection{Further discussion}
\label{Further discussions}


After reviewing the relevant literature, we provide a definition of goal-directedness that is intuitive and more tractable to compute than existing definitions, and a method to create a goal-directedness classifier. We first show that within an MDP setting, our definition is measurable and correlates with features associated with power-seeking. Features like a deterministic policy not taking self-loops and maximizing the number of out-arrows visited at each state tend to occur in optimal, power-seeking policies as shown by \cite{turner2023optimal}, and also occur in goal-directed policies according to our methods. We then adapt our methods to RL environments and again provide evidence that our definition is measurable, even in complex environments. 

Finally, we provide a demonstration of our method on LLMs, where we conducted experiments to measure the classifier's ability to identify activations from sparse and dense loss functions, as defined in Sections \ref{equation:sparse} and \ref{equation:dense}. Our preliminary results indicate that it is possible to achieve an accuracy greater than 70 percent for the dense loss function. As shown in Figure \ref{fig:gsm8K_orca}, when a model is fine-tuned on the loss signal from a specific dataset (GSM8K \cite{cobbe2021gsm8k} and the Orca Math dataset \cite{mitra2024orcamath}), and a classifier is trained on a different dataset, the classifier's ability to detect signals within the activations of the dense loss function increases with the training steps. This result is meaningful because it demonstrates that measuring the dense loss function is feasible across different datasets for the same model. In the future, we would want to train a goal-directedness classifier across models trained on multiple different datasets, over different stages and kinds of training, and over different hyperparameters. While this would be more difficult, the classifier would also generalize better. Overall, we believe that the LLM experiments show a promising method to further understand coherence in LLMs. However, these results are not conclusive, and require more experiments to better understand the intricacies of how "goal directedness" appears in activations.

\subsection{Compute disclosure}
\label{Compute disclosure}

For the MDP and RL experiments in Sections 4.1 and 4.2, only a CPU is needed. Note, however, that it takes about three hours on a Thinkpad laptop to run the RL experiments fully; this is reduced significantly by using a more powerful CPU processor.

For the LLM experiments, we used a single A100 SXM with 32 vCPU and 251 GB RAM on \href{https://www.runpod.io/}{Runpod}.

\subsection{Credit to existing packages}

In addition to the citations already present in the paper, we used the following Python (\cite{10.5555/1593511}) packages: PyTorch (\cite{paszke2019pytorch}), NumPy (\cite{Harris_2020}), scikit-learn (\cite{pedregosa2018scikitlearn}), random, tqdm (\cite{casper_da_costa_luis_2024_11107065}), matplotlib (\cite{Hunter:2007}), os, re, math, pickle (\cite{van1995python}), peft (\cite{peft}), transformers (\cite{wolf-etal-2020-transformers}), datasets (\cite{lhoest-etal-2021-datasets}), and wandb (\cite{wandb}).


\end{document}